\documentclass[12pt]{article}
\usepackage[T1]{fontenc}
\usepackage[utf8]{inputenc}
\usepackage{amsmath,amssymb}
\usepackage[hyphens]{url}
\usepackage[sfdefault]{biolinum}
\RequirePackage{color}
\usepackage{enumerate} 
\usepackage{hyperref}
\usepackage{booktabs}
\usepackage{float}
\usepackage{graphicx}
\hypersetup{colorlinks=true,linkcolor=blue,citecolor=blue,filecolor=blue,urlcolor=black}
\usepackage{wrapfig}
\usepackage[ruled,vlined]{algorithm2e}
\SetKwComment{Comment}{/* }{ */}
\SetCommentSty{normal}
\usepackage[letterpaper, marginparsep=1em, margin=1in]{geometry}
\usepackage{doi}
\usepackage{sectsty}
\usepackage{caption}
\usepackage{subcaption}
\usepackage{abstract}

\usepackage{amsthm}%
\usepackage{bbm}

\newtheorem{theorem}{Theorem}%
\newtheorem{proposition}{Proposition}%
\newtheorem{condition}{Condition}

\DeclareMathOperator{\Var}{Var}
\DeclareMathOperator{\E}{\mathbb{E}}

\sectionfont{\sffamily\upshape\large}
\subsectionfont{\sffamily\upshape\normalsize}
\subsubsectionfont{\sffamily\mdseries\upshape\normalsize}
\usepackage[round]{natbib}

\makeatletter
\def\@maketitle{%
	\begin{center}%
		\let \footnote \thanks
		{\Large \@title \par}%
		{\normalsize
			\begin{tabular}[t]{c}%
				\@author
			\end{tabular}\par}%
            {\small \@date}%
	\end{center}%
}
\makeatother

\title{
	\bf \sffamily Persistent Sampling:\\Enhancing the Efficiency of Sequential Monte Carlo 
	\vspace{.25in} 
}
\author{
    \href{https://orcid.org/0000-0001-9489-4612}{\includegraphics[scale=0.08]{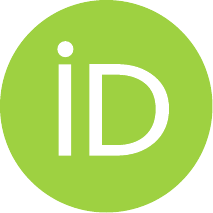}\hspace{1mm}\bf{\sffamily Minas Karamanis}}$^{1,2}$\footnote{mkaramanis@berkeley.edu}
    \and
    \href{https://orcid.org/0000-0003-2262-356X}{\includegraphics[scale=0.08]{orcid.pdf}\hspace{1mm}\bf{\sffamily Uro\v{s} Seljak}}$^{1,2}$
}
\date{
    \vspace{.1in}
    $^1$Berkeley Center for Cosmological Physics, University of California, Berkeley, CA 94720, USA\\%
    $^2$Lawrence Berkeley National Laboratory, 1 Cyclotron Road, Berkeley, CA 94720, USA\\%
    }
\begin{document}\sloppy
	\maketitle
	 \thispagestyle{empty}

\begin{abstract} 
\vspace{-0.15in}
\noindent \emph{Sequential Monte Carlo} (SMC) samplers are powerful tools for Bayesian inference but suffer from high computational costs due to their reliance on large particle ensembles for accurate estimates. We introduce \emph{persistent sampling} (PS), an extension of SMC that systematically retains and reuses particles from all prior iterations to construct a growing, weighted ensemble. By leveraging multiple importance sampling and resampling from a mixture of historical distributions, PS mitigates the need for excessively large particle counts, directly addressing key limitations of SMC such as particle impoverishment and mode collapse. Crucially, PS achieves this without additional likelihood evaluations—weights for persistent particles are computed using cached likelihood values. This framework not only yields more accurate posterior approximations but also produces marginal likelihood estimates with significantly lower variance, enhancing reliability in model comparison. Furthermore, the persistent ensemble enables efficient adaptation of transition kernels by leveraging a larger, decorrelated particle pool. Experiments on high-dimensional Gaussian mixtures, hierarchical models, and non-convex targets demonstrate that PS consistently outperforms standard SMC and related variants, including recycled and waste-free SMC, achieving substantial reductions in mean squared error for posterior expectations and evidence estimates, all at reduced computational cost. PS thus establishes itself as a robust, scalable, and efficient alternative for complex Bayesian inference tasks.

\vspace{0.1in}
\end{abstract}

\section{Introduction}\label{sec1}

\textit{Sequential Monte Carlo} (SMC) samplers have emerged as powerful tools for tackling \textit{Bayesian inference} problems in diverse scientific and engineering fields~\citep{chopin2002sequential, del2006sequential, chopin2020introduction, dai2022invitation}. Their adaptive nature and inherent parallelizability make them attractive for navigating complex, multimodal, and non-linear posterior landscapes~\citep{cappe2007overview}. However, despite their strengths, standard SMC approaches face limitations associated with computational cost and efficiency.

SMC operates by propagating a population of particles through a sequence of intermediate probability distributions, interpolating between a chosen reference distribution (often the prior in Bayesian inference) and the target distribution (often the posterior). This sequence of targets can be constructed by ``data annealing," where observations are gradually introduced (i.e., state-space models), or ``temperature annealing," where the influence of the likelihood function is progressively increased through an effective temperature parameter~\citep{neal2001annealed}. 

To transfer particles from one intermediate distribution to the next, SMC relies on three steps: \emph{reweighting}, \emph{resampling}, and \emph{moving}~\citep{chopin2020introduction}. Reweighting utilizes importance sampling to adjust particle weights based on the agreement between the previous with the current intermediate distribution, guiding the population towards higher probability regions. Resampling then selectively discards low-weight particles and replicates high-weight ones. Finally, the moving step utilizes a Markov kernel to propose transitions for each particle within the current distribution, facilitating the exploration of the current target~\citep{metropolis1953equation}. Notably, the moving step is often the most computationally expensive one, but calculations for each particle are readily parallelizable~\citep{lee2010utility}. Furthermore, as a valuable byproduct, SMC samplers estimate the marginal likelihood (model evidence), crucial for model comparison. 

Despite its advantages, SMC suffers from inherent limitations. These stem from the limited and fixed size of the particle ensemble. The computational cost scales approximately linearly with the number of particles, often prompting a trade-off between accuracy and computational feasibility. Consequently, high-variance estimates plague both marginal likelihoods and posterior distributions. Another challenge lies in post-resampling particle correlation, where many identical particles remain after resampling, necessitating lengthy \textit{Markov chain Monte Carlo} (MCMC) runs for diversification during the moving step.

To address these limitations, we introduce \emph{persistent sampling} (PS), an extension of standard SMC that retains particles from past iterations. This method represents the current target distribution with an growing, weighted collection of particles accumulated over iterations. By treating particles from all previous iterations as approximate samples from a mixture distribution, PS constructs a more diverse and rich sample of the target posterior without additional computational cost. Leveraging information from all iterations rather than just the last one, PS achieves significantly lower variance estimates for the marginal likelihood, enhancing model comparison accuracy. The resampled particles in PS are naturally decorrelated and distinct, reducing the need for extensive MCMC runs for diversification and thereby increasing efficiency. Finally, despite PS requiring more operations per iteration (i.e., recomputing the weights for all historical/persistent particles) compared to SMC, the computational cost (i.e., number of likelihood evaluations) is not increased because the likelihood values entering the weight calculations are already computed and cached in previous iterations. In other words, the recomputation of the weights does not require the evaluation of the likelihood function on any new points and relies only on simple arithmetic operations.  

The idea of using a mixture distribution to perform importance sampling in order to achieve lower-variance estimates was first proposed as the \textit{balance heuristic} in multiple importance sampling (MIS) in the context rendering computer graphics~\citep{veach1995optimally, veach1998robust}. \citet{owen2000safe} showed that the balance heuristic of~\citet{veach1995optimally} matches the deterministic mixture sampling algorithm of~\citet{hesterberg1995weighted}. Although the balance heuristic approach is not unique, it features lower variance compared to alternatives~\citep{elvira2019generalized}. Since then, MIS with a mixture distribution has been utilized to improve the performance of many adaptive importance sampling (AIS) samplers used for general Bayesian computation~\citep{bugallo2017adaptive}. Adaptive multiple importance sampling (AMIS) was the first AIS method to optimally recycle all past simulations using the aforementioned mixture approach~\citep{cornuet2012adaptive}. MIS has also been utilized to improve the sampling efficiency of population Monte Carlo (PMC), a subclass of AIS samplers~\citep{elvira2017improving, elvira2022optimized}. More broadly, the idea of resampling $N$ particles from a much larger pool also appears in the context of waste-free SMC~\citep{dau2022waste}. However, waste-free SMC follows a different approach by resampling $N$ particles from their respective Markov chains instead of past SMC iterations. Finally, in an attempt to utilize the particles of the intermediate distributions to enhance posterior estimates, several recycling techniques have been introduced~\citep{nguyen2014improving}. In this context, recycling acts as a post-processing step after sampling with SMC is completed, where particles from all intermediate distributions are reweighted to target the posterior. 

The rest of this paper is organized as follows: Section \ref{sec:background} provides the necessary background, including a detailed description of the standard SMC algorithm and recycling post-processing procedures. Section \ref{sec:method} introduces the method following the construction of the PS algorithm. Section \ref{sec:experiments} presents numerical results, demonstrating the superior empirical performance of PS. Finally, Section \ref{sec:discussion} offers a summary and discussion of the main results and potential extensions of the algorithm.

\section{Background}
\label{sec:background}

Bayesian inference relies on Bayes' theorem, which mathematically expresses how a subjective degree of belief, or \textit{prior probability distribution} $\pi(\theta)\equiv P(\theta\vert \mathcal{M})$, is updated given data, $D$, to form the \textit{posterior probability distribution} $\mathcal{P}(\theta) \equiv P(\theta|D, \mathcal{M})$~\citep{jaynes2003probability,mackay2003information}. Here, $\mathcal{M}$ is the model with parameters $\theta$. This theorem is captured by the equation 
\begin{equation}
    \label{eq:bayes_rule}
    \mathcal{P}(\theta) = \frac{\mathcal{L}(\theta)\pi(\theta)}{\mathcal{Z}}
\end{equation}
where $\mathcal{L}(\theta)\equiv P(D|\theta,\mathcal{M})$ is the \textit{likelihood function}, representing the probability of observing the data as a function of the parameters $\theta$, and $\mathcal{Z}\equiv P(D\vert\mathcal{M})$ is the \textit{model evidence},\textit{marginal likelihood}, or simply the \textit{normalizing constant}, that is, the probability of observing the data under the model. The strength of Bayesian inference lies in its formal framework for incorporating prior knowledge and systematically updating this knowledge in light of new data. Typically, the likelihood, $\mathcal{L}(\theta)$, and the prior, $\pi(\theta)$, are known and one seeks to estimate the posterior, $\mathcal{P}(\theta)$, and the evidence, $\mathcal{Z}$.

\subsection{Sequential Monte Carlo}
\label{sec:smc}

SMC propagates a set of $N$ particles, $\lbrace \theta_{t}^{i} \rbrace_{i=1}^{N}$, through a sequence of probability distributions, $p_{t}(\theta)$ for $t=1\,,\dots\,,T$. This is achieved through the repeated application of a series of reweighting, resampling, and moving steps that are described in detail below. The SMC pseudocode is presented in detail in Algorithm \ref{alg:smc}.

The sequence of probability distributions, $p_{t}(\theta)$ for $t=1\,,\dots\,,T$, is generally chosen such that $p_{1}(\theta)$ corresponds to a reference distribution from which sampling is straightforward, and $p_{T}(\theta)$ corresponds to the generally intractable target distribution that we aim to approximate. In the context of Bayesian inference, we typically focus on a \emph{temperature annealing} sequence of the form
\begin{equation}
    \label{eq:annealing_sequence}
    p_{t}(\theta) = \frac{\mathcal{L}^{\beta_{t}}(\theta)\pi(\theta)}{\mathcal{Z}_{t}}
\end{equation}
that interpolates between the prior $p_{1}(\theta)=\pi(\theta)$ and posterior distribution $p_{T}(\theta)=\mathcal{P}(\theta)=\mathcal{L}(\theta)\pi(\theta)/\mathcal{Z}_{T}$ given a sequence of temperature values $0=\beta_{1}<\dots<\beta_{t}<\dots<\beta_{T}=1$~\citep{neal2001annealed}. The choice of Eq. \ref{eq:annealing_sequence} is neither unique nor necessarily optimal for all Bayesian inference tasks. However, this choice offers both concreteness in terms of presentation and, most importantly, many computational simplifications. Alternative options include geometric paths between successive partial posteriors $p_{t}(\theta)=P(D_{1:t}|\theta, \mathcal{M})$ for sequential data assimilation, or truncated distributions $p_{t}(\theta) \propto \pi(\theta)\mathbbm{1}\lbrace \mathcal{L}(\theta)\geq \ell_{t}\rbrace$~\citep{dai2022invitation}. 

\subsubsection{Reweighting}

The reweighting step involves using importance sampling (IS) to weigh particles drawn from the previous distribution, $p_{t-1}(\theta)$, to target the current distribution, $p_{t}(\theta)$~\citep{del2006sequential}. The weights are defined as
\begin{equation}
    \label{eq:importance_weights}
    W_{t}^{i} \equiv \frac{p_{t}(\theta_{t-1}^{i})}{p_{t-1}(\theta_{t-1}^{i})} = \frac{\hat{\mathcal{Z}}_{t-1}}{\hat{\mathcal{Z}}_{t}}w_{t}^{i}\,,
\end{equation}
where $w_{t}^{i}$ are the unnormalized weights given by
\begin{equation}
    \label{eq:unnormalized_importance_weights}
    w_{t}^{i} = \mathcal{L}(\theta_{t-1}^{i})^{\beta_{t}-\beta_{t-1}}\,,
\end{equation}
which follows directly from Eq. \ref{eq:annealing_sequence}. It is worth noting that the particular form of the weights of Eq. \ref{eq:importance_weights} is a special case of a more general framework, one that encompasses both forward and backward kernels~\citep{dai2022invitation}. In the general formulation, the weights accommodate auxiliary transition kernels—often referred to as the forward kernel and its associated backward counterpart—which allow for additional flexibility and improved performance in SMC samplers. Essentially, these kernels help correct for the bias introduced by the proposal distribution used in the particle update and can be tailored to incorporate more complex dependencies or adaptive strategies. The specific form presented above corresponds to the situation where the backward kernel is chosen conveniently so that the weight update simplifies to the expression in the first part of Eq. \ref{eq:importance_weights}. For the purposes of our paper, this simplified form of the weights suffices.

Assuming sufficient overlap between $p_{t-1}(\theta)$ and $p_{t}(\theta)$, the weighted set of particles $\lbrace \theta_{t-1}^{i}, W_{t}^{i}\rbrace_{i=1}^{N}$ should be distributed according to $p_{t}(\theta)$. The unknown ratio of normalizing constants in Eq. \ref{eq:importance_weights} can be estimated as
\begin{equation}
    \label{eq:normalizing_constant_ratio}
    \frac{\hat{\mathcal{Z}}_{t}}{\hat{\mathcal{Z}}_{t-1}} = \frac{1}{N}\sum_{i=1}^{N}w_{t}^{i}\,,
\end{equation}
where the weights $w_{t}^{i}$ are given by Eq. \ref{eq:unnormalized_importance_weights}. In the context of Bayesian inference, the reference distribution is chosen to be the prior $\pi(\theta)$ with $\mathcal{Z}_{1}=1$, further simplifying calculations. In this case, $\mathcal{Z}_{T}$ corresponds to the model evidence, $\mathcal{Z}=p(d\vert\mathcal{M})$ (i.e., marginal likelihood), for a model, $\mathcal{M}$, and is crucial in the task of \textit{Bayesian model comparison}. 

In order to ensure the low variance of the weights $\lbrace W_{t}^{i}\rbrace_{i=1}^{N}$ and the expected values constructed based on them, the temperature level $\beta_{t}$ is typically determined adaptively. This is achieved by approximately maintaining a specified \textit{effective sample size} (ESS), estimated as
\begin{equation}
    \label{eq:effective_sample_size}
    \hat{\textrm{ESS}}_{t} = \frac{(\sum_{i=1}^{N}w_{t}^{i})^{2}}{\sum_{i=1}^{N}(w_{t}^{i})^{2}} = \frac{1}{\sum_{i=1}^{N}(W_{t}^{i})^{2}} \,.
\end{equation}
$\hat{\mathrm{ESS}}$ is maximized when the weight distribution is uniform, in which case $\hat{\mathrm{ESS}}=N$~\citep{kong1994sequential, jasra2011inference}. More formally, $\beta_{t}$ is determined as
\begin{equation}
    \label{eq:next_beta}
    \beta_{t} = \inf_{\beta\in[\beta_{t-1},1]}\left\lbrace\hat{\textrm{ESS}}_{t} \geq \alpha N\right\rbrace\,,
\end{equation}
where values of $\alpha$ typically range from $0.5$ to $0.999$, depending on the specific requirements of the inference task. Eq. \ref{eq:next_beta} is commonly solved numerically using the bisection method~\citep{arfken2011mathematical}. Higher values of $\alpha$ lead to a finer sequence of temperature levels and generally more robust results, albeit more computationally expensive. SMC typically terminates when $\beta_{t}$ becomes 1, corresponding to the target posterior distribution. Using Eq. \ref{eq:next_beta} to determine $\beta_{t}$ is a form of adaptation and can incur a level of bias in the final results. Despite this, the introduced bias is often negligible and the estimates remain consistent~\citep{delmoral2012adaptive,beskos2016convergence}, rendering the use of Eq. \ref{eq:next_beta} a standard practice in the field. It should be mentioned that while the adaptation of $\beta_{t}$ using the ESS is a popular choice, its estimators are not always reliable, in which case an alternative definition can be considered~\citep{elvira2022rethinking}. 

\subsubsection{Resampling}

Following the reweighting step, the particles, $\lbrace \theta_{t-1}^{i}\rbrace_{i=1}^{N}$, are resampled with probabilities proportional to their weights, $\lbrace W_{t}^{i}\rbrace_{i=1}^{N}$ (i.e., multinomial resampling). The aim of this procedure is to eliminate particles with low weights and replicate the ones with high weights. Different resampling methods, such as multinomial and systematic resampling, with varying theoretical characteristics have been proposed to perform this step~\citep{li2015resampling, gerber2019negative}.

\subsubsection{Moving}

In SMC, after resampling, the particles undergo diversification through a sequence of MCMC steps to explore the distribution $p_{t}(\theta)$. The choice of MCMC kernel, $\mathcal{K}_{t}$, varies based on application-specific needs. For high-dimensional scenarios, gradient-based kernels like \emph{Metropolis-adjusted Langevin algorithm} (MALA) or \emph{Hamiltonian Monte Carlo} (HMC) are effective~\citep{grenander1994representations,rossky1978brownian,roberts1996exponential,roberts1998optimal,duane1987hybrid,neal2011mcmc}. In contrast, low-to-moderate dimensional, non-differentiable problems may benefit from gradient-free approaches such as \emph{random-walk Metropolis} (RWM) or \emph{slice sampling} (SS)~\citep{metropolis1953equation, neal2003slice}. \emph{Independence Metropolis-Hastings} (IMH) kernels are also useful in certain low-dimensional cases~\citep{south2019sequential}. Additionally, machine learning-assisted methods like preconditioned MCMC have shown promise for highly correlated targets~\citep{karamanis2022accelerating}.

While the list of potential algorithms is not exhaustive, a key advantage of SMC is leveraging the empirical particle distribution to create effective MCMC proposals, contrasting with methods like \emph{parallel tempering} (PT) that lack such auxiliary information~\citep{swendsen1986replica,geyer1991computing}. One common strategy involves using the empirical covariance matrix of particles for RWM or as the mass matrix in MALA and HMC~\citep{chopin2002sequential}. Examples of advanced applications include using a copula-based model calibrated on particle distribution for IMH, or employing normalizing flows trained on particle distribution for preconditioned MCMC~\citep{south2019sequential, karamanis2022accelerating}. Furthermore, the MCMC kernel's hyperparameters, such as the proposal scale or mass matrix, can be fine-tuned using the mean Metropolis acceptance rate from the previous iteration.

\begin{algorithm}[hbt!]
\caption{Sequential Monte Carlo}\label{alg:smc}

\KwData{Number of particles $N$, ESS fraction $\alpha$, prior density $\pi(\theta)$, likelihood function $\mathcal{L}(\theta)$, Markov kernel $\mathcal{K}_{t}$ with invariant density $p_{t}(\theta)$}
\KwResult{Samples $\lbrace \theta^{i}_{T} \rbrace_{i=1}^{N}$ from $\mathcal{P}(\theta)=\mathcal{L}(\theta)\pi(\theta)/\mathcal{Z}$ and marginal likelihood estimate $\hat{\mathcal{Z}}\equiv\hat{\mathcal{Z}}_{T}$}
$t \gets 1,\;\beta_{t} \gets 0,\,\hat{\mathcal{Z}}_{1}\gets 1$\;
\For{$i\gets 1$ \KwTo $N$}{
    $\theta_{1}^{i}\sim \pi(\theta)$ \Comment*[r]{Sample prior distribution}
    }
\While{$\beta_{t} < 1$}{
    $t \gets t + 1$\;
    $\beta_{t}\gets \inf_{\beta\in[\beta_{t-1},1]} \left\lbrace \hat{\textrm{ESS}}_{t} \geq \alpha N \right\rbrace$ \Comment*[r]{Update temperature}
    \For{$i\gets 1$ \KwTo $N$}{
        $w_{t}^{i}\gets \mathcal{L}(\theta_{t-1}^{i})^{\beta_{t}-\beta_{t-1}}$ \Comment*[r]{Compute unnormalized importance weights}
        }
    $\hat{\mathcal{Z}}_{t}\gets \hat{\mathcal{Z}}_{t-1}\times N^{-1}\sum_{i=1}^{N}w_{t}^{i}$ \Comment*[r]{Update marginal likelihood}
    \For{$i\gets 1$ \KwTo $N$}{
        $W_{t}^{i}\gets w_{t}^{i}\times \hat{\mathcal{Z}}_{t}/\hat{\mathcal{Z}}_{t-1}$ \Comment*[r]{Normalize importance weights}
        }
    $\lbrace \tilde{\theta}_{t}^{i}\rbrace_{i=1}^{N}\sim \texttt{resample}(\lbrace \theta_{t-1}^{i}\rbrace_{i=1}^{N}, \lbrace W_{t}^{i}\rbrace_{i=1}^{N})$ \Comment*[r]{Resample particles}
    \For{$i\gets 1$ \KwTo $N$}{
        $\theta_{t}^{i}\gets \mathcal{K}_{t}(\tilde{\theta}_{t}^{i})$  \Comment*[r]{ Propagate particles}
        }
    }

\end{algorithm}

\subsection{Computing Posterior Expectation Values}

\subsubsection{Standard Approach}

The canonical approach of using the output of SMC to compute expectation values is to use final population of particles. The estimate of a posterior expectation value, that is,
\begin{equation}
    \label{eq:expectation_value_target}
    \mathbb{E}_{\mathcal{P}}[f] = \int_{\Theta}f(\theta)\mathcal{P}(\theta)d\theta
\end{equation}
is given by
\begin{equation}
    \label{eq:expectation_value_canonical_estimate}
    \Hat{f} = \frac{1}{N}\sum_{i=1}^{N}f(\theta_{T}^{i})
\end{equation}

\subsubsection{Recycling}
\label{sec:recycling}

When it comes to the estimation of posterior expectation values, the standard approach can be wasteful as it only considers the final positions of the particles, $\theta_{T}^{i}$, and discards the previous generations of particles, $\theta_{t}^{i}$ for $t=1,\,\dots,\,T-1$, that do not target the posterior directly. To remedy this uneconomical use of computation, several recycling schemes have been proposed that aim to reweight samples drawn from $p_{t}(\theta)$ with $t\neq T$ to target $p_{T}(\theta)$.

\citet{gramacy2010importance} proposed a technique to combine multiple estimates $\Hat{f}_{t}$ into a single one given by
\begin{equation}
    \label{eq:gramacy_combined_estimator}
    \Hat{f} = \sum_{t=1}^{T}\lambda_{t}\Hat{f}_{t}
\end{equation}
where $\lambda_{t}=\hat{\textrm{ESS}}_{t}'/\sum_{t=1}^{T}\hat{\textrm{ESS}}_{t}'$ is the normalized $\hat{\textrm{ESS}}$ of the particles $\theta_{t}^{i}$ reweighted to target the posterior. The motivation behind this method is that estimators that have low $\hat{\textrm{ESS}}$ will generally contribute less to the final estimate than those with high $\hat{\textrm{ESS}}$. Although originally proposed in the context of importance tempering, \citet{nguyen2014improving} and \citet{finke2015extended} extended this approach to SMC.

An alternative approach for combining samples drawn from multiple distributions $p_{t}(\theta)$ is to assume that the total collection of samples are drawn from the mixture of distributions $p_{t}(\theta)$, where the mixture weights represent the proportion of samples coming from each distribution~\citep{veach1995optimally, owen2000safe, elvira2019generalized}. In the case of equal proportions, the importance density is defined as:
\begin{equation}
    \label{eq:mixture_importance_density}
    \Tilde{p}(\theta) = \frac{1}{T}\sum_{t=1}^{T}p_{t}(\theta)
\end{equation}
where the densities $p_{t}(\theta)$ need to be properly normalized. Using Eq. \ref{eq:mixture_importance_density} as the importance density, the importance weights for particles of all iterations are:
\begin{equation}
    \label{eq:mixture_importance_weights}
    \Tilde{w}_{t}^{i} = \frac{\mathcal{L}(\theta_{t}^{i})}{\frac{1}{T}\sum_{t'=1}^{T}\mathcal{L}(\theta_{t}^{i})^{\beta_{t'}}\hat{\mathcal{Z}}_{t}^{-1}}
\end{equation}
for $i=1\,,\dots\,,N$ and $t=1\,,\dots\,,T$. Since the normalization constants $\hat{\mathcal{Z}}_{t}$ are generally unknown, one could use their SMC estimates given by Eq. \ref{eq:normalizing_constant_ratio}. Additionally, one could use the weights of Eq. \ref{eq:mixture_importance_weights} to estimate the normalizing constant $\hat{\mathcal{Z}}_{T}$. However, as shown elsewhere and further verified in Section \ref{sec:experiments}, this yields only marginally better estimates at the expense of a more complicated estimator~\citep{south2019sequential}.

\citet{nguyen2014improving} compared the ESS-based recycling approach to the mixture-based one and deduced that the latter performs marginally better despite relying on estimates of the normalizing constant $\hat{\mathcal{Z}}_{t}$. Our own experiments verified this conclusion and indicated that the noisy estimates of $\hat{\mathcal{Z}}_{t}$ are less problematic than the noisy estimates of the $\lambda_{t}$ factors in Eq. \ref{eq:gramacy_combined_estimator}. In our tests, the latter typically required \textit{ad hoc} thresholds to perform well. For these reasons, we will use the mixture-based approach of recycling in our numerical experiments in Section \ref{sec:experiments}. We will refer to SMC with recycling as \emph{recycled sequential Monte Carlo} (RSMC).

\section{Method}
\label{sec:method}

\subsection{Persistent Sampling}

\begin{figure}[ht!]
    \centering
    \includegraphics[width=15.5cm]{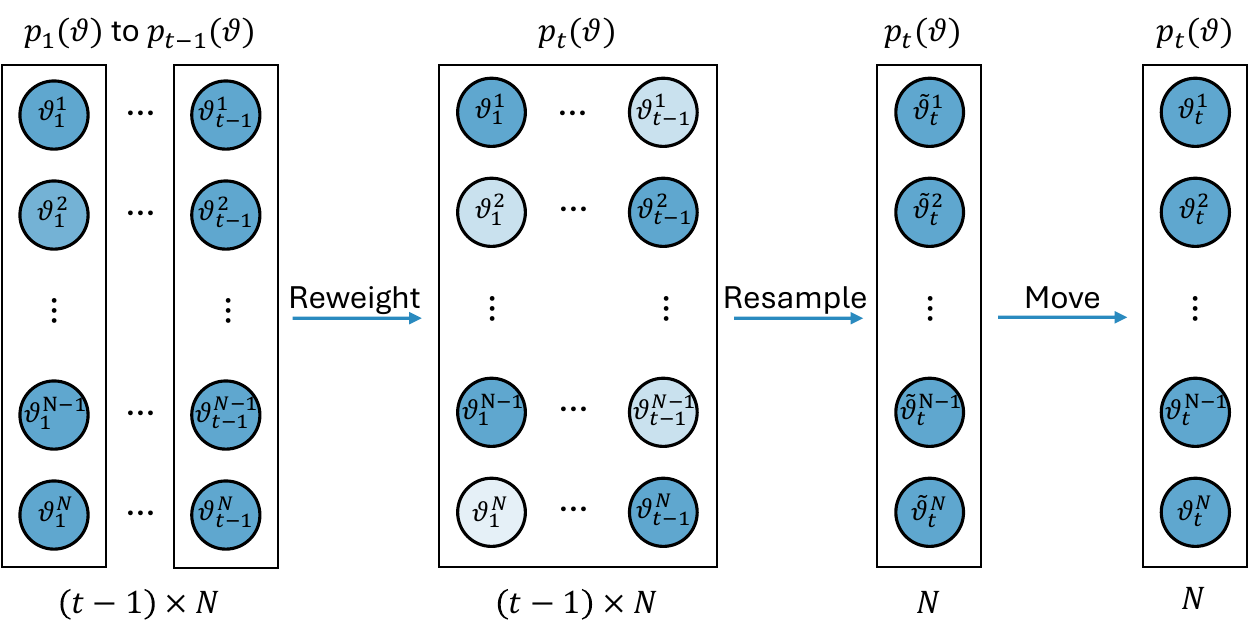}
    \caption{Illustration of the reweighting, resampling, and moving steps in PS. The $(t-1)\times N$ particles representing $p_{1}(\theta)$ to $p_{t-1}(\theta)$ are reweighted to target $p_{t}(\theta)$. Out of these $(t-1)\times N$ particles, $N$ are resampled and moved to better approximate $p_{t}(\theta)$.}
    \label{fig:rules}
\end{figure}

Similar to SMC, PS traverses a sequence of probability distributions, $p_{t}(\theta)$ for $t=1\,,\dots\,,T$, through a series of reweighting, resampling, and moving steps, as illustrated in Fig. \ref{fig:rules}. The main difference is that the current collection of particles, $\lbrace \theta_{t}^{i}\rbrace_{i=1}^{M}$, at any iteration does not simply originate from the previous distribution, $p_{t-1}(\theta)$, by means of the aforementioned three operations. In contrast, particles are reweighted and resampled from all previous iterations, not just the last one. In other words, the particle approximation of $p_{t-1}(\theta)$ is determined directly by the particle approximations of $p_{s}(\theta)$ for $s=1,\,\dots\,,t-1$. The PS pseudocode is presented in detail in Algorithm \ref{alg:ps}.

\subsubsection{Reweighting}

In the reweighting step, PS assigns a single weight $W_{tt'}^{i}$ to each particle $\theta_{t'}^{i}$, where $t$ and $t'<t$ denote the current and past iteration, respectively. The aim is that the weighted particles will target $p_{t}(\theta)$ despite being drawn from different distributions. Similar to multiple importance sampling~\citep{veach1995optimally, veach1998robust}, we can treat $\theta_{t'}^{i}$ for $t'=1\,,\dots\,,t-1$ and $i=1,\,\dots\,,N$ as approximate samples from the mixture distribution
\begin{equation}
    \label{eq:mixture_distribution}
    \Tilde{p}_{t}(\theta) = \frac{1}{t-1}\sum_{s=1}^{t-1}p_{s}(\theta)\,,
\end{equation}
where $p_{s}(\theta)$ is the usual annealed target at iteration $s$. Employing Eq. \ref{eq:mixture_distribution} as the importance density, we can derive the following weights:
\begin{equation}
    \label{eq:persistent_weights}
    W_{tt'}^{i} = \frac{p_{t}(\theta_{t'}^{i})}{\frac{1}{t-1}\sum_{s=1}^{t-1}p_{s}(\theta_{t'}^{i})} = \frac{w_{tt'}^{i}}{\hat{\mathcal{Z}}_{t}}\,,
\end{equation}
where the $w_{tt'}^{i}$ are the unnormalized weights given by
\begin{equation}
    \label{eq:unnormalized_persistent_weights}
    w_{tt'}^{i} = \frac{\mathcal{L}(\theta_{t'}^{i})^{\beta_{t}}}{\frac{1}{t-1}\sum_{s=1}^{t-1}\mathcal{L}(\theta_{t'}^{i})^{\beta_{s}}\hat{\mathcal{Z}}_{s}^{-1}}
\end{equation}
for $t'=1\,,\dots\,,t-1$ and $i=1\,,\dots\,,N$. The normalization constant $\hat{\mathcal{Z}}_{t}$ can be estimated as
\begin{equation}
    \label{eq:normalization_cosntant}
    \hat{\mathcal{Z}}_{t} = \frac{1}{t-1}\sum_{t'=1}^{t-1}\left( \frac{1}{N}\sum_{i=1}^{N} w_{tt'}^{i} \right)
\end{equation}
It should be noted that evaluation of the weights of Eqs. \ref{eq:persistent_weights} and \ref{eq:unnormalized_importance_weights} do not require new additional evaluations of the likelihood function. Instead, the values $\mathcal{L}(\theta_{t'}^{i})$ are already known since they have been computed at iteration $t' < t$. In other words, the computational complexity of PS in terms of likelihood evaluations is the same as that of standard SMC.

However, unlike SMC, the $\hat{\textrm{ESS}}$ in PS is computed using the persistent weights. Since the persistent particles are generally more numerous than the standard particles of SMC, the $\hat{\textrm{ESS}}$ in PS can, in principle, be many times greater than in SMC. The $\hat{\textrm{ESS}}$ of the persistent particles is defined as:
\begin{equation}
    \label{eq:peristent_effective_sample_size}
        \hat{\textrm{ESS}}_{t} = \frac{\left[\sum_{t'=1}^{t-1}(\sum_{i=1}^{N}w_{tt'}^{i})\right]^{2}}{\sum_{t'=1}^{t-1}\left[\sum_{i=1}^{N}(w_{tt'}^{i})^{2}\right]} = \frac{1}{\sum_{t'=1}^{t-1}\left[\sum_{i=1}^{N}(W_{tt'}^{i})^{2}\right]}
\end{equation}

In standard SMC, the $\hat{\textrm{ESS}}_{t}$ can be approximately linked to the inverse of the $\chi^{2}$ divergence between $p_{t-1}(\theta)$ and $p_{t}(\theta)$~\citep{elvira2022rethinking}. The relation quantifies the trade-off between distributional similarity and weight stability in SMC. It formalizes why gradual transitions in terms of $\chi^{2}$ divergence are critical for maintaining sample quality, while abrupt changes degrade performance. This intuition naturally extends to PS as well, where the $\hat{\textrm{ESS}}_{t}$ is now related to the $\chi^{2}$ divergence between the mixture density $\tilde{p}_{t}(\theta)$ and $p_{t}(\theta)$.

PS determines the next temperature level, $\beta_{t}$, adaptively by maintaining a constant $\hat{\textrm{ESS}}$ throughout the run, similarly to SMC. To find the next $\beta_{t}$, one needs to solve Eq. \ref{eq:next_beta} numerically. However, $\alpha$ is no longer restricted to be strictly smaller than 1, and can take values in the interval $(0, t-1)$. When $\alpha \geq 1$, in the initial $\lfloor\alpha\rfloor + 1$ iterations of PS, $\beta_{t}$ will repeatedly be set to 0 and $p_{t}(\theta)$ will remain the prior distribution $\pi(\theta)$. This provides the option to reduce the computational cost of these initial iterations by performing independent sampling from the prior instead of MCMC transitions. After iteration $\lfloor\alpha\rfloor + 1$, the value of $\beta$ will start increasing as usual, annealing the distribution from the prior toward the posterior.

PS typically terminates when $\beta_{t}$ becomes 1, corresponding to the target posterior distribution. However, an additional advantage of PS is that one can continue sampling even after $\beta_{t}=1$ until a prespecified ESS is gathered. In this scenario, $\beta_{t}=\beta_{t+1}=\dots=\beta_{T-1}=\beta_{T}=1$. This can provide more samples from the posterior distribution without going through all the $T$ iterations from the beginning.

\subsubsection{Resampling}

Resampling of particles in PS can be performed using the same techniques as in SMC (i.e., multinomial resampling, systematic resampling, etc.). However, the difference is that in PS one generally resamples $N$ new particles from $M\equiv (t-1)\times N \geq N$ persistent particles. In other words, the pool of available candidates grows throughout the run but the number of resampled particles remains the same. This is, at least partially, the reason for the improved performance of PS compared to SMC. \emph{Propagation of chaos} theory suggests that in the limit where $N\ll (t-1)\times N$, the $N$ resampled particles behave essentially as independent samples from the probability distribution~\citep{moral2004feynman}. This means that the $N$ resampled particles in PS are less correlated than the respective $N$ resampled particles in SMC. This well established result is briefly formalized in the following Proposition.

\medskip

\begin{proposition}[Propagation of chaos]
    Let $\{(X_i, W_i)\}_{i=1}^M$ be a system of $M$ weighted particles with normalized weights $W_i \geq 0$ and $\sum_{i=1}^M W_i = 1$, and let $\nu^M = \sum_{i=1}^M W_i \delta_{X_i}$ denote the associated empirical measure. Suppose that as $M \to \infty$, $\nu^M$ converges weakly to a probability measure $\nu$. For any fixed integer $N \geq 1$, let $Y_1, \dots, Y_N$ be obtained by resampling $N$ times with replacement from $\{(X_i, W_i)\}_{i=1}^M$. Then, as $M \to \infty$, the joint distribution of $(Y_1, \dots, Y_N)$ converges weakly to $\nu^{\otimes N}$. Consequently, the resampled particles $\{Y_j\}_{j=1}^N$ are asymptotically independent and identically distributed (i.i.d.) with common law $\nu$.
\end{proposition}

\medskip

\noindent \textbf{Interpretation:} This result formalizes the propagation of chaos principle in the context of resampling: when $M \gg N$, dependencies among the resampled particles vanish as $M \to \infty$, yielding approximately i.i.d. samples from the limiting measure $\nu$. The convergence of $\nu^M$ to $\nu$ ensures that the weighted empirical distribution approximates the target measure, and resampling from it inherits this asymptotic independence.

\subsubsection{Moving}

Unlike the reweighting and resampling steps of PS that constitute clear extensions of the respective SMC components, the moving step is largely unaltered. In other words, the same MCMC methods can be used to diversify the particles following the resampling steps. That said, the moving step in PS boasts a major advantage -- the adaptation of the MCMC kernel can be performed by utilizing the more numerous persistent particles. This naturally leads to kernels that better adapt to the geometrical characteristics of the target. For instance, in the case where a covariance matrix needs to be adapted, typically at least $N\geq 2\times D$ particles are required to get a non-singular estimate. In high-dimensions this can become prohibitively expensive for standard SMC. In contrast, the same covariance matrix estimation in PS is performed utilizing $(t-1)\times N\geq N$ samples. Furthermore, even in cases where the number of SMC particles are sufficient, PS will still lead to better estimates given the greater size of its ensemble of persistent particles. The same principles apply to the estimation of components of more complicated kernels (e.g., mixture distributions, normalizing flows, etc.).

\begin{algorithm}[hbt!]
\caption{Persistent Sampling}\label{alg:ps}

\KwData{Number of particles $N$, ESS fraction $\alpha$, prior density $\pi(\theta)$, likelihood function $\mathcal{L}(\theta)$, Markov kernel $\mathcal{K}_{t}$ with invariant density $p_{t}(\theta)$}
\KwResult{Samples $\lbrace \theta^{i}_{T} \rbrace_{i=1}^{N}$ from $\mathcal{P}(\theta)=\mathcal{L}(\theta)\pi(\theta)/\mathcal{Z}$ and marginal likelihood estimate $\mathcal{Z}\equiv\mathcal{Z}_{T}$}
$t \gets 1,\;\beta_{t} \gets 0,\,\hat{\mathcal{Z}}_{1}\gets 1$\;
\For{$i\gets 1$ \KwTo $N$}{
    $\theta_{1}^{i}\sim \pi(\theta)$ \Comment*[r]{Sample prior distribution}
    }
\While{$\beta_{t} < 1$}{
    $t \gets t + 1$\;
    $\beta_{t}\gets \inf_{\beta\in[\beta_{t-1},1]} \left\lbrace \hat{\textrm{ESS}}_{t} \geq \alpha N \right\rbrace$ \Comment*[r]{Update temperature}
    \For{$t'\gets 1$ \KwTo $t$}{
        \For{$i\gets 1$ \KwTo $N$}{
            $w_{tt'}^{i} \gets \frac{\mathcal{L}(\theta_{t'}^{i})^{\beta_{t}}}{\frac{1}{t-1}\sum_{s=1}^{t-1}\mathcal{L}(\theta_{t'}^{i})^{\beta_{s}}\hat{\mathcal{Z}}_{s}^{-1}}$ \Comment*[r]{Compute unnormalized importance weights}
        }
    }
    $\hat{\mathcal{Z}}_{t} \gets \frac{1}{t-1}\sum_{t'=1}^{t-1}\left( \frac{1}{N}\sum_{i=1}^{N} w_{tt'}^{i} \right)$ \Comment*[r]{Update marginal likelihood}
    \For{$t'\gets 1$ \KwTo $t$}{
        \For{$i\gets 1$ \KwTo $N$}{
            $W_{tt'}^{i}\gets \frac{w_{tt'}^{i}}{\hat{\mathcal{Z}}_{t}}$ \Comment*[r]{Normalize importance weights}
        }
    }
    $\lbrace \tilde{\theta}_{t}^{i}\rbrace_{i=1}^{N}\sim \texttt{resample}(\lbrace\lbrace \theta_{t'}^{i}\rbrace_{i=1}^{N}\rbrace_{t'=1}^{t-1}, \lbrace\lbrace W_{tt'}^{i}\rbrace_{i=1}^{N}\rbrace_{t'=1}^{t-1})$ \Comment*[r]{Resample particles}
    \For{$i\gets 1$ \KwTo $N$}{
        $\theta_{t}^{i}\gets \mathcal{K}_{t}(\tilde{\theta}_{t}^{i})$  \Comment*[r]{ Propagate particles}
        }
    }

\end{algorithm}

\subsection{Computing Expectations}

Computing posterior expectation values in PS can be done by
utilizing the persistent particles with their associated normalized weights. This leads to estimators of the form:
\begin{equation}
    \label{eq:weighted_estimate}
    \Hat{f} = \sum_{t'=1}^{T} \sum_{i=1}^{N}W_{Tt'}^{i}f(\theta_{t'}^{i})
\end{equation}
where $T$ is the total number of iterations. Of course one could alternatively resample the persistent particles $\theta_{t'}^{i}$ with weights $W_{Tt'}^{i}$ at the end of the run and use the resampled particles $\Tilde{\theta}_{T}^{i}$ to compute expectations. However, the use of the weighted persistent particles should be preferred when possible since the resampling procedure can introduce some variance.

\subsection{Bias and Consistency}

A celebrated property of standard SMC methods is the unbiasedness of their normalizing constant estimates. However, it is important to note that even in standard SMC, this unbiasedness is not always realized in practice, particularly when adaptive temperature schedules are employed. In contrast to standard SMC, PS, while offering significant variance reduction, introduces bias into its estimates of both normalizing constants and posterior expectations.  However, as we demonstrate in the theorems presented in this section, both the normalizing constant and posterior expectation estimators in PS are consistent.  Furthermore, we provide a theoretical quantification of the bias in the normalizing constant estimates, detailing its scaling behavior with respect to the number of particles and iterations. This theoretical analysis, in conjunction with the extensive empirical evaluations in the subsequent section (proofs are provided in Appendix A), underscores and validates a well-established principle in statistical literature: that accepting a potentially negligible bias can often be a highly beneficial trade-off, especially when it leads to a substantial and practically relevant reduction in variance.

\subsubsection{Consistency of Normalizing Constant Estimates}

\begin{theorem}[Consistency of Normalizing Constant Estimates]
Under technical conditions \ref{C1}-\ref{C4} (specified below), for any fixed $t$:
\begin{equation}
    \hat{\mathcal{Z}}_t \xrightarrow{p} \mathcal{Z}_t \text{ as } N \to \infty
\end{equation}
\end{theorem}

\begin{condition}[Likelihood ratio boundedness]\label{C1}
For all $t, s \leq T$, there exists $M_{ts} < \infty$ such that:
\begin{equation}
\sup_{\theta \in \Theta} \frac{\mathcal{L}(\theta)^{\beta_t}}{\mathcal{L}(\theta)^{\beta_s}} \leq M_{ts}.
\end{equation}
This ensures the importance weights $w_{tt'}^i$ remain bounded across iterations.
\end{condition}

\begin{condition}[Normalizing constant positivity]\label{C2}
There exists $\epsilon > 0$ such that for all $t \leq T$:
\begin{equation}
\mathcal{Z}_t = \int \mathcal{L}(\theta)^{\beta_t} \pi(\theta) d\theta \geq \epsilon.
\end{equation}
This guarantees that the denominator in weight calculations is bounded away from zero.
\end{condition}

\begin{condition}[MCMC kernel regularity]\label{C3}
For each $t \leq T$, the MCMC kernel $\mathcal{K}_t$ satisfies:
\begin{equation}
    \sup_{\theta \in \Theta} \|\mathcal{K}_t^n(\theta, \cdot) - p_t(\cdot)\|_{TV} \leq C_t \rho_t^n,
\end{equation}
for constants $C_t < \infty$, $\rho_t \in (0, 1)$, and all $n \geq 1$. This ensures geometric ergodicity of the mutation step.
\end{condition}

\begin{condition}[Temperature schedule regularity]\label{C4}
There exists $\delta > 0$ such that:
\begin{equation}
    \min_{2 \leq t \leq T} (\beta_t - \beta_{t-1}) \geq \delta.
\end{equation}
This condition prevents arbitrarily small annealing steps, which could potentially destabilize the algorithm.
\end{condition}

\subsubsection{Consistency of Posterior Expectations}

\begin{theorem}[Consistency of Posterior Expectations]
Under Conditions \ref{C1}-\ref{C5}, for any bounded function $f \in L^\infty(\Theta)$, the PS estimator satisfies:
\begin{equation}
    \hat{\mu}_N(f) = \sum_{t'=1}^T \sum_{i=1}^N W_{Tt'}^i f(\theta_i^{t'}) \xrightarrow{p} \mathbb{E}_\mathcal{P}[f(\theta)] \quad \text{as } N \to \infty.
\end{equation}
\end{theorem}

\begin{condition}[Function regularity]\label{C5}
For test functions $f: \Theta \to \mathbb{R}$, assume:
\begin{equation}
   \|f\|_\infty = \sup_{\theta \in \Theta} |f(\theta)| < \infty. 
\end{equation}
This is required for controlling Monte Carlo errors in expectation estimates.
\end{condition}

\subsubsection{Bias Magnitude}

\begin{theorem}[Bias Magnitude]
Under Conditions \ref{C1}-\ref{C4}, for finite $N$ and $t\ge3$, the relative bias of the normalizing constant estimator satisfies:
\begin{equation}
\left|\frac{\E[\hat{\mathcal{Z}}_t]}{\mathcal{Z}_t} - 1\right| \le \frac{C_t}{N} + \mathcal{O}\left(\frac{1}{N^2}\right),
\end{equation}
where $C_t$ depends on the temperature schedule $\{\beta_s\}_{s=1}^t$, likelihood ratio bounds, and the number of iterations $t$.
\end{theorem}

\section{Experiments}
\label{sec:experiments}

In this section, we conduct a cost-matched comparison of four Sequential Monte Carlo (SMC) variants:

\begin{itemize}
    \item \textbf{Persistent Sampling (PS):} Our proposed method (Section~\ref{sec:method}, Algorithm~\ref{alg:ps}) estimates posterior expectations via Eq.~\ref{eq:weighted_estimate} and computes the log normalizing constant using Eq.~\ref{eq:normalization_cosntant}.
    
    \item \textbf{Sequential Monte Carlo (SMC):} The canonical implementation (Section~\ref{sec:smc}, Algorithm~\ref{alg:smc}) estimates posterior expectations with Eq.~\ref{eq:expectation_value_canonical_estimate} and iteratively updates the log normalizing constant via Eq.~\ref{eq:normalizing_constant_ratio}.
    
    \item \textbf{Recycled SMC (RSMC):} This variant applies a post-processing recycling step to reweight all particles toward the posterior density (Section~\ref{sec:recycling}). While it shares the log normalizing constant estimator (Eq.~\ref{eq:normalizing_constant_ratio}) and posterior expectation estimator (Eq.~\ref{eq:expectation_value_canonical_estimate}) with standard SMC, RSMC employs the mixture importance weights defined in Equation~\ref{eq:mixture_importance_weights}.
    
    \item \textbf{Waste-Free SMC (WFSMC)}~\citep{dau2022waste}\textbf{:} Similar to PS, WFSMC enhances particle diversity by resampling $N$ particles from an expanded pool of $M > N$ weighted particles. After $k$ MCMC propagation steps, the subsequent generation combines $M = k \times N$ states generated during the MCMC evolution into the next generation of particles.
\end{itemize}

To ensure fair comparison, we match computational costs by controlling the number of likelihood evaluations. All methods use identical particle counts ($N$), MCMC steps, and an optimally tuned Random-Walk Metropolis sampler ($23.4\%$ average acceptance rate, proposal covariance adapted via Robbins-Monro diminishing adaptation). For SMC and RSMC, we fix the effective sample size (ESS) threshold at $\alpha = 90\%$. We then calibrate the ESS thresholds of PS and WFSMC to achieve computational cost parity (within $<1\%$ error in likelihood evaluations). While WFSMC requires only minor adjustments to $\alpha$, PS achieves comparable costs with $\alpha = 300\%$, regardless of $N$ or MCMC steps. We test particle counts $N \in [32, 256]$ and MCMC steps $k \in [25, 400]$.

Typically, one cares about the bias and variance of the first and second posterior moments, and the log marginal likelihood $\log\mathcal{Z}$ estimate. Generally, the MSE of an estimator can be decomposed as the sum of its variance plus the square of its bias. For this reason, MSE is a very informative metric for comparing different samplers. We evaluate estimator quality using mean squared error (MSE) for:  
\begin{enumerate}
    \item The log marginal likelihood $\log\mathcal{Z}$
    \item First ($f_1(\theta) = \theta$) and second ($f_2(\theta) = \theta^2$) posterior moments  
\end{enumerate}

For $\log\mathcal{Z}$, we compute MSE across $L=200$ independent runs as:
\begin{equation}
    \label{eq:mse_logz}
    \textrm{MSE}_{\log\mathcal{Z}} = \frac{1}{L}\sum_{\ell=1}^{L}\left( \log\hat{\mathcal{Z}}_{\ell}-\log\mathcal{Z}\right)^{2},
\end{equation}
where $\log\mathcal{Z}$ is a high-accuracy reference value obtained from $200$ 16,384-particle SMC runs with $\alpha=99.9\%$.  

For the $D$-dimensional posterior moments, we use the \textit{maximum squared bias across dimensions}:
\begin{equation}
    \label{eq:squared_bias}
    b_{i}^{2} = \underset{d}{\max}\left( \frac{\hat{f}_{i,d}-\mu(f_{i,d})}{\sigma(f_{i,d})}\right)^{2},
\end{equation}
where $\hat{f}_{i,d}$ is the coordinate-wise mean across $L=200$ runs, and $\mu(f_{i,d})$, $\sigma(f_{i,d})$ denote posterior moments estimated from the reference SMC runs. This metric is essentially an MSE formulation applied to posterior expectations. By normalizing the bias in each coordinate with the corresponding posterior standard deviation, it reduces the multivariate output into a single, scale-invariant number which highlights the worst-case error across dimensions. This metric is a well-established and common approach in sampler comparisons \citep{hoffman2022tuning, grumitt2022deterministic, robnik2023microcanonical}.

Before presenting the detailed results for each benchmark, we provide a high-level summary of the key findings from our cost-matched comparisons. Across all four challenging test cases—a multimodal Gaussian mixture, a non-convex Rosenbrock function, a high-dimensional sparse logistic regression model, and a Bayesian hierarchical model with a funnel-shaped posterior —PS consistently and substantially outperforms standard SMC, RSMC, and WFSMC. This superiority is evident in all evaluated metrics: the MSE of the log marginal likelihood ($\log \mathcal{Z}$) and the maximum squared bias for both first and second posterior moments ($b_{1}^{2}$, $b_{2}^{2}$). The performance advantage of PS is particularly pronounced in the most complex scenarios. For instance, in the Bayesian hierarchical model, other methods struggle to reduce error even with a high number of MCMC steps, whereas PS shows rapid error reduction. Similarly, for the Gaussian mixture target, PS is far more effective at correctly estimating the relative weight of the two modes, a task where the other methods show significant bias. While WFSMC provides some improvement over standard SMC, it remains less accurate than PS, and RSMC offers only marginal gains, primarily in posterior moment estimation at high particle counts. The following subsections provide a detailed, quantitative analysis of these findings.

\subsection{Gaussian Mixture}

\begin{figure}[ht!]
    \centering
    \includegraphics[width=7.5cm,trim={0.5cm 0.5cm 0.75cm 0.5cm},clip]{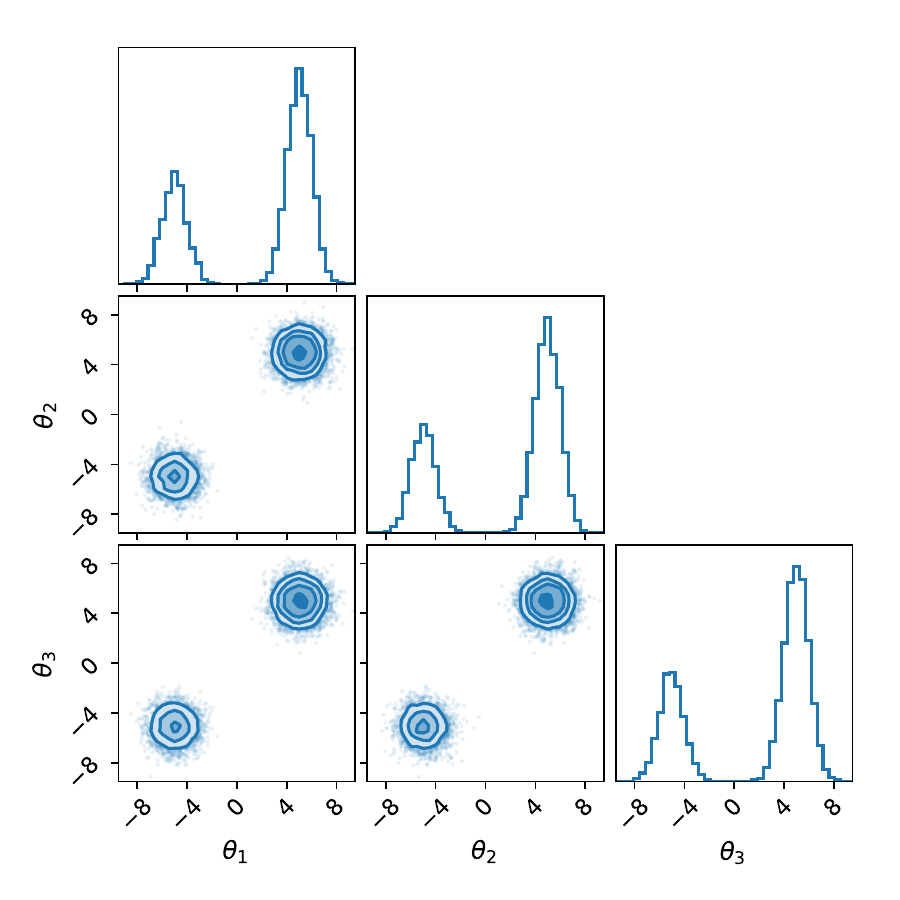}
    \caption{1D and 2D marginal posterior distributions of the first three parameters of the Gaussian mixture target.}
    \label{fig:bimodal}
\end{figure}

A challenging benchmark problem for evaluating the performance of SMC algorithms is sampling from a multimodal distribution with well-separated modes. We consider a 16-dimensional bimodal  \textit{Gaussian mixture} model with a likelihood function:
\begin{equation}
    \label{eq:bimodal}
    \mathcal{L}(\boldsymbol{\theta})=\frac{1}{3}\mathcal{N}(\boldsymbol{\theta} \vert -5,\mathbf{I}) + \frac{2}{3}\mathcal{N}(\boldsymbol{\theta} \vert 5, \mathbf{I})\,,
\end{equation}
where $\mathcal{N}(\boldsymbol{\theta}|\boldsymbol{\mu}, \mathbf{\Sigma})$ denotes the probability density function of a normal variable $\boldsymbol{\theta}$ with mean $\boldsymbol{\mu}$ and covariance matrix $\mathbf{\Sigma}$. We chose a multivariate uniform distribution $\pi(\boldsymbol{\theta})=\mathcal{U}(\boldsymbol{\theta}\vert -10,10)$ as the prior to cover both modes.
This bimodal target features well-separated and unequally weighted modes, making it difficult for samplers to efficiently explore the two modes and accurately estimate properties like the normalizing constant and the relative weight of the two modes. Fig. \ref{fig:bimodal} illustrates the 1D and 2D marginal posterior contours for the first three parameters of this target.

\begin{figure}[ht!]
    \centering
    \includegraphics[width=16.0cm,trim={0 0.20cm 0 0.20cm},clip]{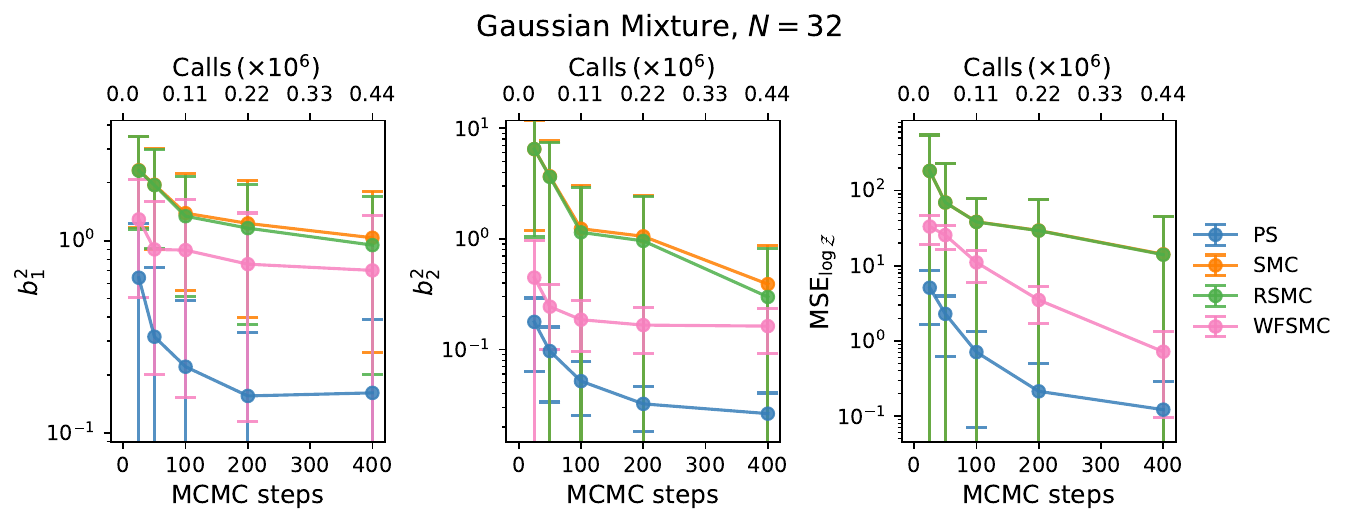}
    \vskip 0.5cm
    \includegraphics[width=16.0cm,trim={0 0.20cm 0 0.20cm},clip]{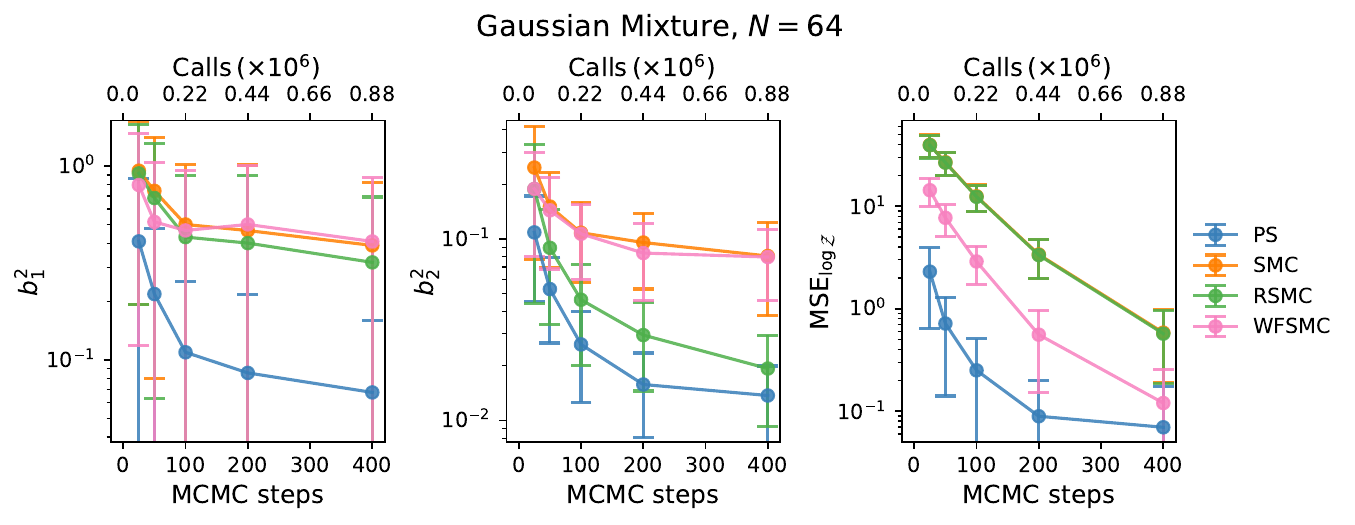}
    \vskip 0.5cm
    \includegraphics[width=16.0cm,trim={0 0.20cm 0 0.20cm},clip]{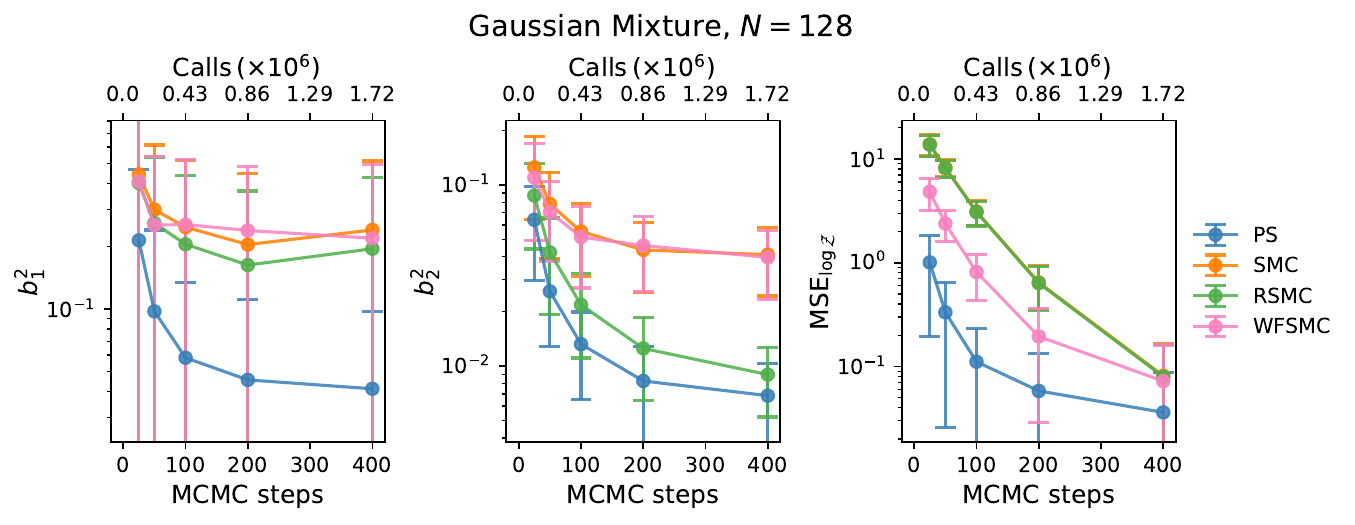}
    \caption{Sampling performance for the first (\emph{left}) and second (\emph{middle}) posterior moments and log marginal likelihood ($\textrm{MSE}_{\log \mathcal{Z}}$; \emph{right}) for the 16-dimensional Gaussian mixture target using $N=32$ (\textit{top}), $N=64$ (\textit{middle}), and $N=128$ (\textit{bottom}) particles.}
    \label{fig:bimodal_comp}
\end{figure}

Our cost-matched comparison across four SMC variants reveals distinct performance patterns in the 16-dimensional Gaussian mixture benchmark, as shown in Fig. \ref{fig:bimodal_comp}. PS consistently outperforms competing methods, achieving the lowest MSE for the log marginal likelihood ($\log\mathcal{Z}$) and the smallest maximum squared bias ($b_1^2$, $b_2^2$) for posterior moments across all tested particle counts ($N \in [32, 128]$) and MCMC steps ($k \in [25, 400]$). RSMC demonstrates modest improvements over standard SMC in posterior moment estimation, particularly for $b_2^2$ at larger particle counts ($N=64$ and $N=128$), where it surpasses WFSMC but remains less accurate than PS. However, RSMC shows no discernible advantage over SMC in $\log\mathcal{Z}$ estimation. WFSMC achieves intermediate performance: while it reduces $\log\mathcal{Z}$ MSE compared to SMC and RSMC, it lags behind PS. For posterior moments, WFSMC outperforms SMC and RSMC with $N=32$ but matches SMC/RSMC in $b_1^2$ and underperforms RSMC in $b_2^2$ at higher $N$. All methods exhibit gradual error reduction as MCMC steps increase, though PS maintains a consistent and substantial efficiency advantage, particularly in mitigating bias for the unequally weighted, multimodal target. As demonstrated by the large $b_1^2$ values of SMC, RSMC, and WFSMC, these methods struggle to capture the correct weight balance between the two modes of the posterior, even when using a large number of particles. This underscores PS’s robustness in navigating complex posterior geometries while maintaining computational parity with other methods.

\subsection{Rosenbrock Function}

\begin{figure}[ht!]
    \centering
    \includegraphics[width=7.5cm,trim={0.5cm 0.5cm 0.5cm 0.5cm},clip]{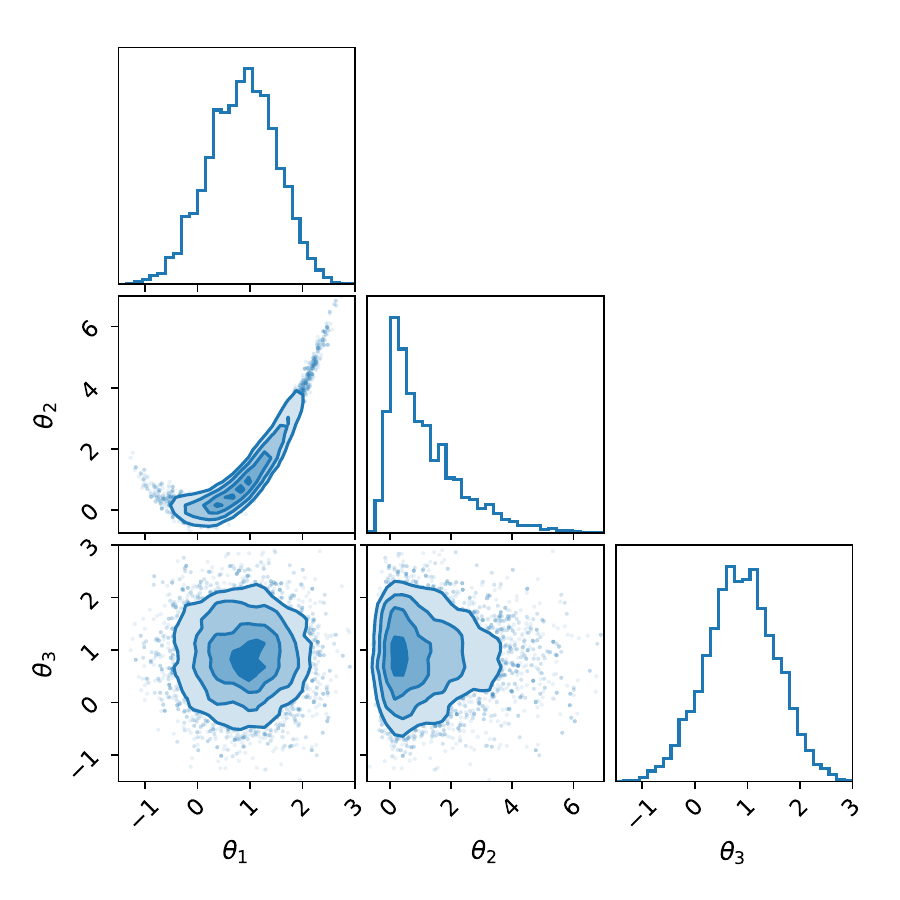}
    \caption{1D and 2D marginal posterior distributions of the first three parameters of the Rosenbrock target.}
    \label{fig:rosenbrock}
\end{figure}

The \textit{Rosenbrock function} is a well known non-convex optimization benchmark~\citep{rosenbrock1960automatic}. The function is particularly challenging for optimization algorithms due to its narrow, curved valley containing the minimum. Here we consider a $16$-dimensional extension of this function as the log-likelihood function given by
\begin{equation}
    \label{eq:rosenbrock}
        \log\mathcal{L}(\boldsymbol{\theta}) = -\sum_{i=1}^{8}\left[ 10 (\theta_{2i-1}^{2}-\theta_{2i})^{2}\right. \left.+(\theta_{2i-1}-1)^{2}\right]
\end{equation}
with a multivariate normal distribution $\pi(\boldsymbol{\theta})=\mathcal{N}(\boldsymbol{\theta}\vert 0, 5^{2}\,\mathbf{I})$ as the prior distribution. Fig. \ref{fig:rosenbrock} shows the 1D and 2D marginal posterior contours for the first three parameters of this target.

\begin{figure}[ht!]
    \centering
    \includegraphics[width=16.0cm,trim={0 0.20cm 0 0.20cm},clip]{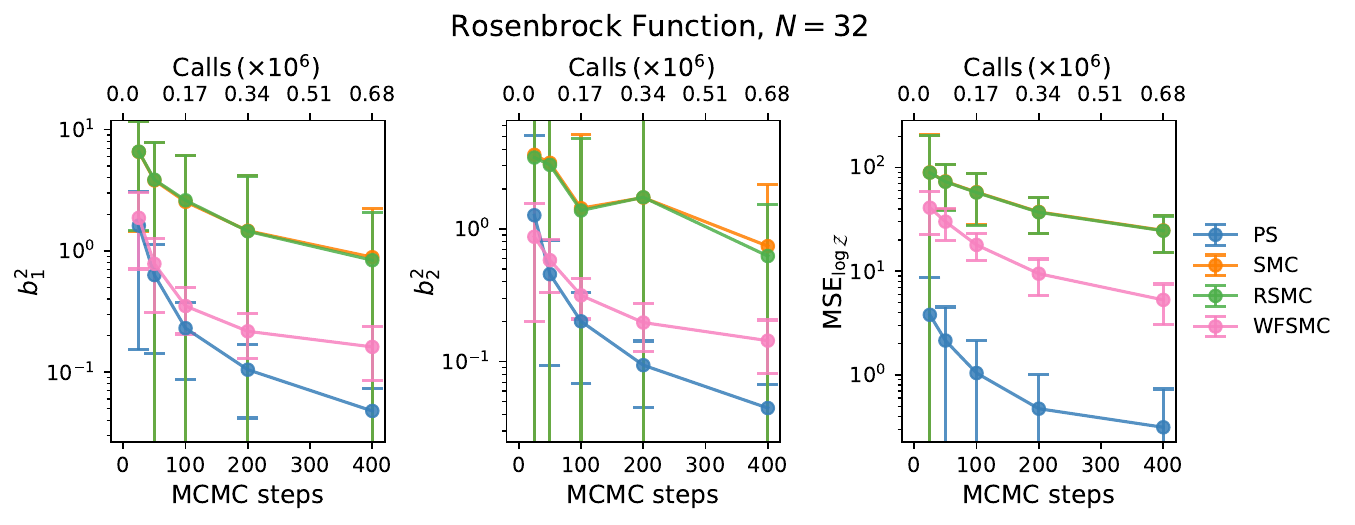}
    \vskip 0.5cm
    \includegraphics[width=16.0cm,trim={0 0.20cm 0 0.20cm},clip]{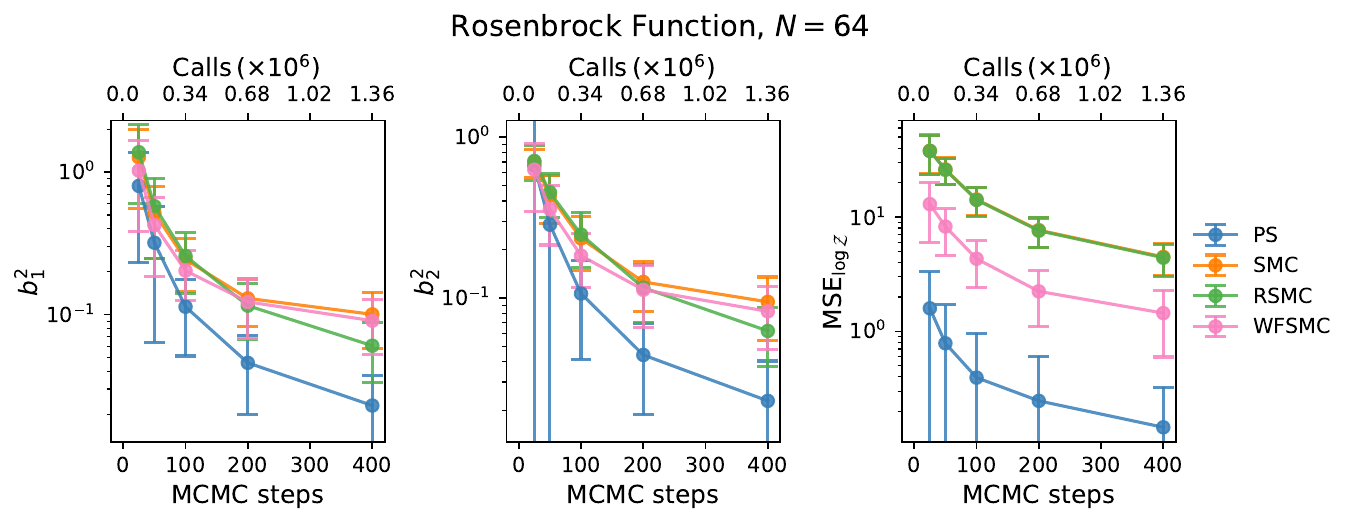}
    \vskip 0.5cm
    \includegraphics[width=16.0cm,trim={0 0.20cm 0 0.20cm},clip]{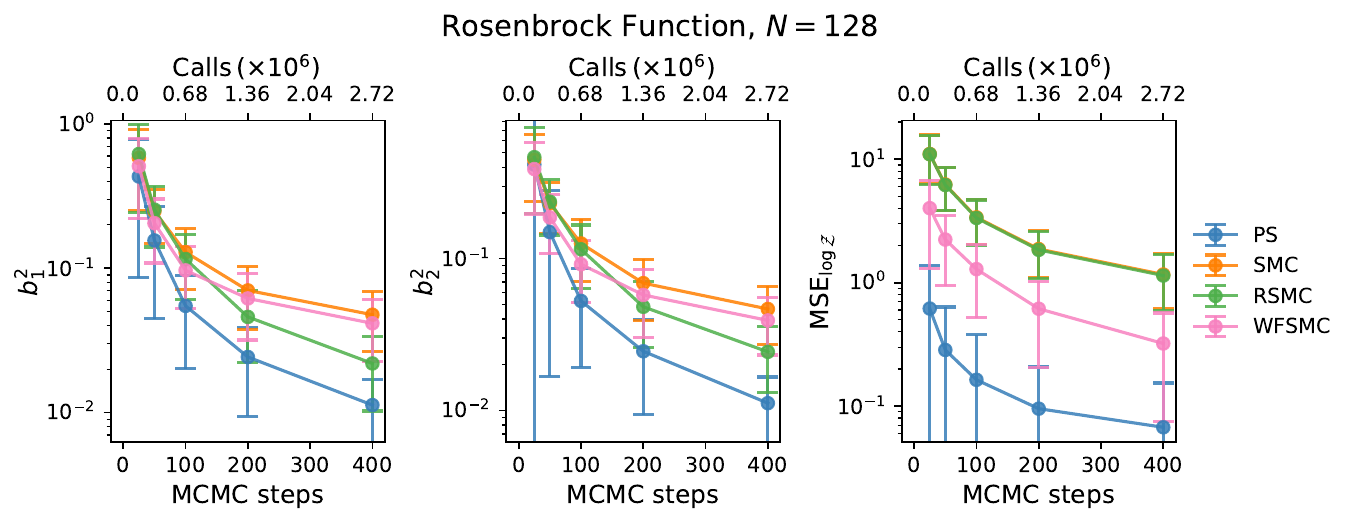}
    \caption{Sampling performance for the first (\emph{left}) and second (\emph{middle}) posterior moments and log marginal likelihood ($\textrm{MSE}_{\log \mathcal{Z}}$; \emph{right}) for the 16-dimensional Rosenbrock function target using $N=32$ (\textit{top}), $N=64$ (\textit{middle}), and $N=128$ (\textit{bottom}) particles.}
    \label{fig:rosen_comp}
\end{figure}

The 16-dimensional Rosenbrock benchmark highlights distinct performance patterns across methods under its challenging non-Gaussian geometry, as shown in Fig. \ref{fig:rosen_comp}. PS consistently achieves the lowest maximum squared bias ($b_1^2$, $b_2^2$) and log marginal likelihood MSE ($\textrm{MSE}_{\log\mathcal{Z}}$) across all particle counts ($N \in [32, 128]$) and MCMC steps ($k \in [25, 400]$). While PS and WFSMC perform comparably in $b_1^2$ and $b_2^2$ at low MCMC steps ($k < 100$), PS maintains a decisive advantage in $\textrm{MSE}_{\log\mathcal{Z}}$, even in this regime. RSMC mirrors standard SMC in most configurations, with minor deviations at large $k$: for $N=64$ and $N=128$ with $k=400$, RSMC slightly surpasses both SMC and WFSMC in posterior moment accuracy, though it remains less precise than PS. All methods exhibit gradual error reduction as MCMC steps increase, but PS’s superiority in $b_1^2$, $b_2^2$, and $\textrm{MSE}_{\log\mathcal{Z}}$ grows more pronounced at higher $k$. For $\textrm{MSE}_{\log\mathcal{Z}}$, PS outperforms WFSMC, which itself surpasses SMC and RSMC—the latter two showing indistinguishable performance. At $N=32$, WFSMC reduces $b_1^2$ and $b_2^2$ relative to SMC/RSMC but remains less accurate than PS. This gap narrows at $N=64$ and $N=128$, where WFSMC’s posterior moment performance aligns with SMC. These trends underscore PS’s robustness in navigating the Rosenbrock function’s curved, narrow valleys, even under matched computational costs.

\subsection{Sparse Logistic Regression}

In this experiment, we explore a \textit{sparse logistic regression} model in 51 dimensions, applied to the German credit dataset~\citep{dua2017uci}. This dataset comprises 1,000 data points, each representing an individual who has borrowed from a financial institution. The dataset classifies these individuals into two categories: those deemed as good credit risks and those considered bad credit risks, according to the bank's evaluation criteria. Each data point contains 24 numerical covariates, which may or may not serve as useful predictors for determining credit risk. These covariates include factors such as age, gender, and savings information. The model incorporates a hierarchical structure and utilizes a horseshoe prior for the logistic regression parameters~\citep{carvalho2009handling}. The horseshoe prior promotes sparsity in the regression coefficients, effectively performing variable selection.

\begin{figure}[ht!]
    \centering
    \includegraphics[width=7.5cm,trim={0.5cm 0.5cm 0.5cm 0.5cm},clip]{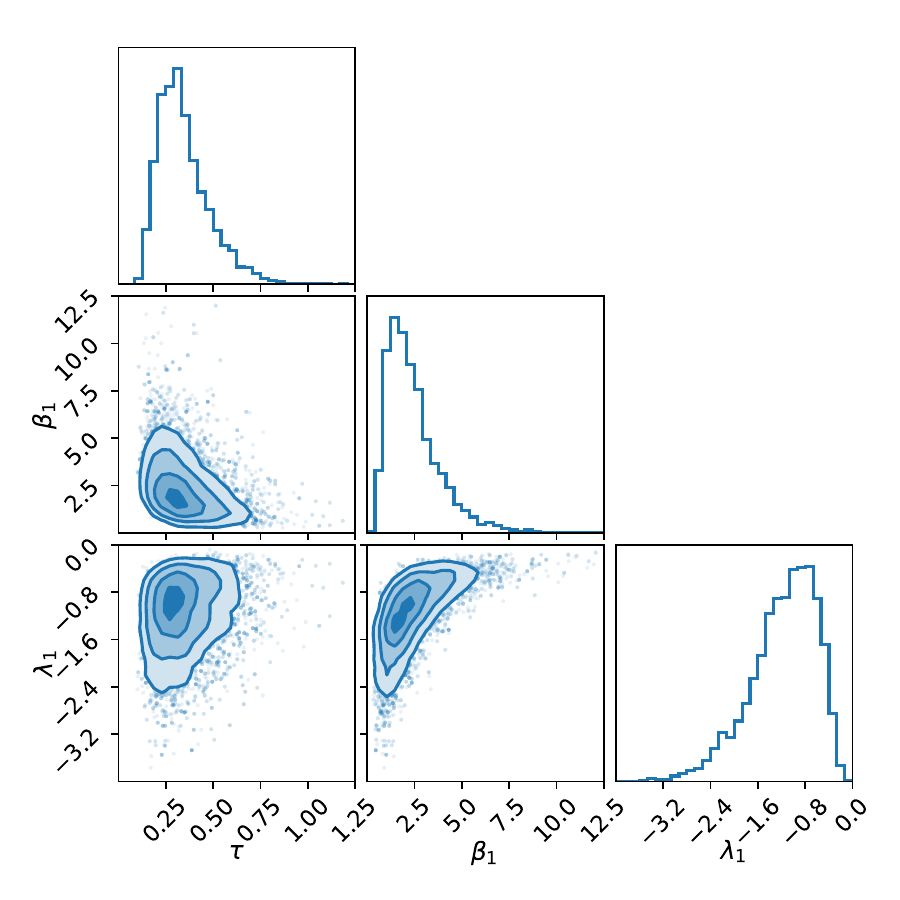}
    \caption{1D and 2D marginal posterior distributions of the $\tau$, $\beta_{1}$, and $\lambda_{1}$ parameters of the sparse logistic regression with the German credit data.}
    \label{fig:gc}
\end{figure}

We define the likelihood function as follows:
\begin{equation}
\mathcal{L}(\mathbf{y} | \boldsymbol{\beta}, \boldsymbol{\lambda}, \tau) = \prod_{i=1}^{1000} \text{Bernoulli}\left(y_{i} | \sigma((\tau\boldsymbol{\lambda} \odot \boldsymbol{\beta})^{T}X_{i})\right),
\end{equation}
where $\sigma(\cdot)$ denotes the sigmoid function. $\boldsymbol{\beta}$ and $\boldsymbol{\lambda}$ are $25$-dimensional vectors and $\tau$ is a scalar quantity. The priors for the model parameters are given by:
\begin{equation}
    \pi(\boldsymbol{\beta},\boldsymbol{\lambda},\tau) = \text{Gamma}(\tau | 1/2, 1/2) \times\prod_{j=1}^{25} \mathcal{N}(\beta_{j} | 0, 1) \text{Gamma}(\lambda_{j} | 1/2, 1/2).
\end{equation}
Fig. \ref{fig:gc} shows the 1D and 2D marginal posterior contours for the $\tau$, $\beta_{1}$, and $\lambda_{1}$ parameters of this target, emphasizing the highly non-Gaussian nature of this target originating by the hierarchical structure of the model.

\begin{figure}[ht!]
    \centering
    \includegraphics[width=16.0cm,trim={0 0.20cm 0 0.20cm},clip]{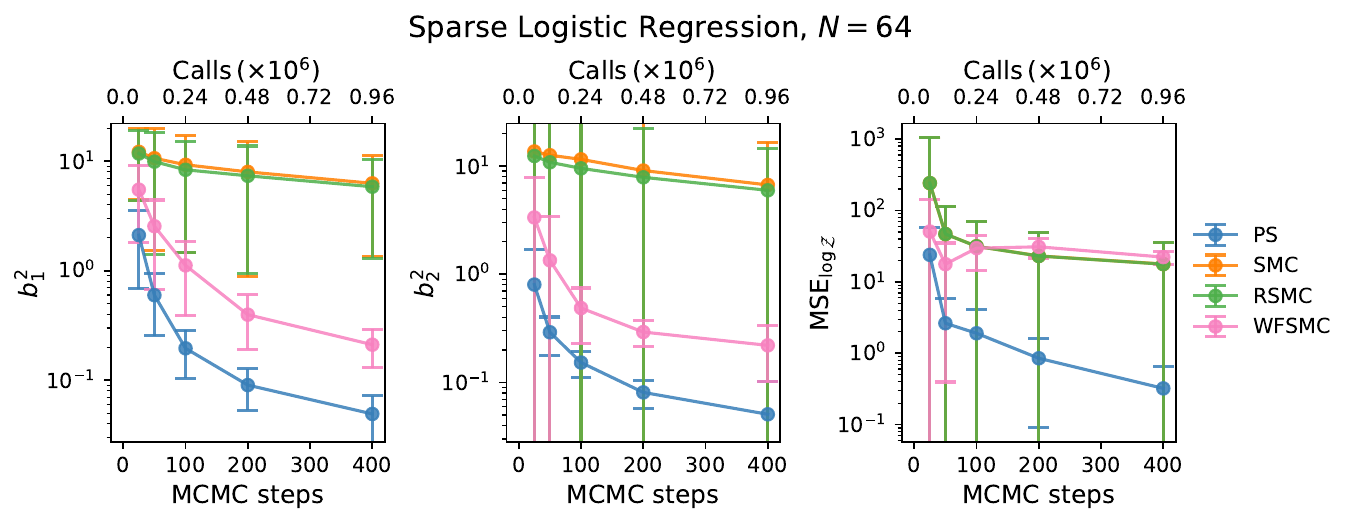}
    \vskip 0.5cm
    \includegraphics[width=16.0cm,trim={0 0.20cm 0 0.20cm},clip]{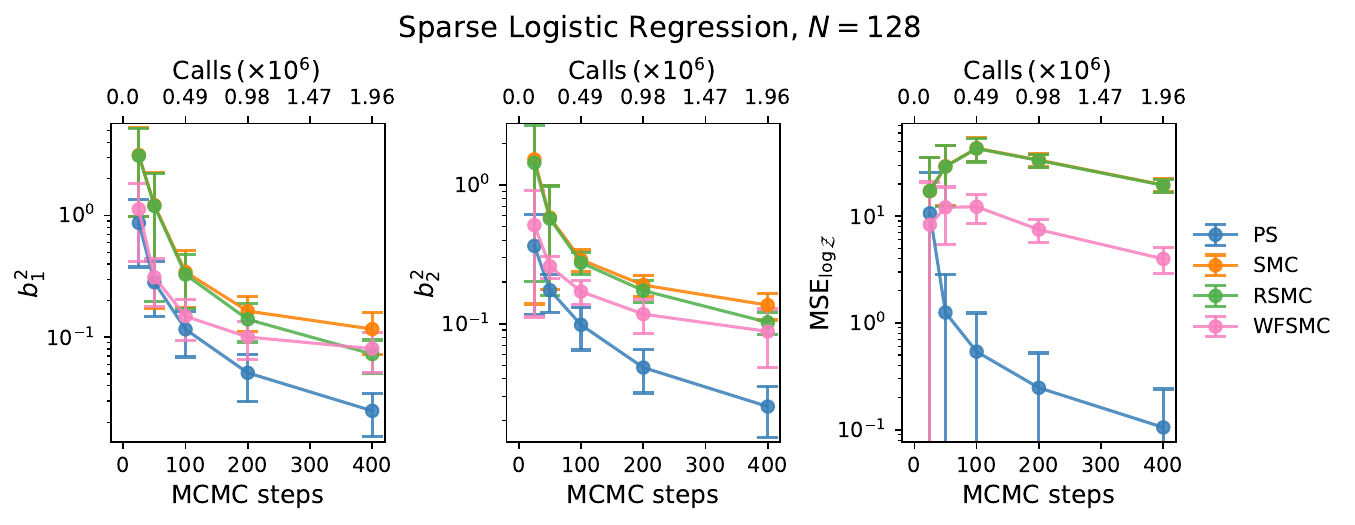}
    \vskip 0.5cm
    \includegraphics[width=16.0cm,trim={0 0.20cm 0 0.20cm},clip]{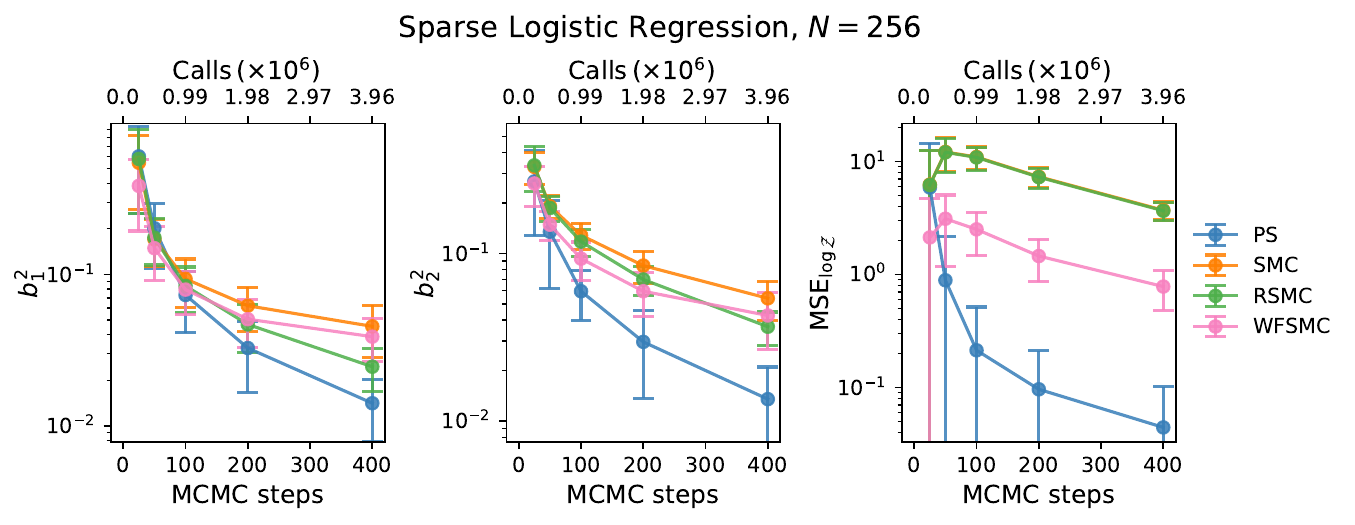}
    \caption{Sampling performance for the first (\emph{left}) and second (\emph{middle}) posterior moments and log marginal likelihood ($\textrm{MSE}_{\log \mathcal{Z}}$; \emph{right}) for the 51-dimensional German Credit Sparse Logistic Regression target using $N=64$ (\textit{top}), $N=128$ (\textit{middle}), and $N=256$ (\textit{bottom}) particles.}
    \label{fig:germ_comp}
\end{figure}

The 51-dimensional sparse logistic regression benchmark underscores PS’s consistent high performance across all metrics, even in this high-dimensional, non-Gaussian hierarchical setting, as shown in Fig. \ref{fig:germ_comp}. PS achieves the lowest maximum squared bias ($b_1^2$, $b_2^2$) and log marginal likelihood MSE ($\textrm{MSE}_{\log\mathcal{Z}}$) for all particle counts ($N \in [64, 256]$) and MCMC steps ($k \in [25, 400]$). RSMC performs nearly identically to standard SMC in most cases, including $\textrm{MSE}_{\log\mathcal{Z}}$, though RSMC marginally surpasses SMC in $b_1^2$ and $b_2^2$ at $k=400$ for $N=128$ and $N=256$. WFSMC exhibits variable performance: for $N=64$, it significantly outperforms SMC/RSMC in posterior moment accuracy but remains less precise than PS. At $N=128$, WFSMC slightly exceeds SMC/RSMC for $k < 200$ but matches them at $k=400$; for $N=256$, it aligns with SMC/RSMC across all $k$. Notably, PS and WFSMC show comparable $b_1^2$ and $b_2^2$ at low MCMC steps ($k \leq 100$) for $N=128$ and $N=256$, though PS retains superiority in $\textrm{MSE}_{\log\mathcal{Z}}$. All methods struggle with $\textrm{MSE}_{\log\mathcal{Z}}$ at $k=25$, reflecting the target’s inherent difficulty. While SMC, RSMC, and WFSMC exhibit only gradual $\textrm{MSE}_{\log\mathcal{Z}}$ improvements even at $k=400$, PS achieves rapid error reduction, attaining far lower values. These results highlight PS’s robustness in high-dimensional, sparsity-promoting models, where efficient exploration of hierarchical parameter spaces is critical.  

\subsection{Bayesian Hierarchical Model}

\textit{Bayesian hierarchical modeling} (BHM) plays a critical role in contemporary science, integrating multiple sub-models within a unified framework~\citep{gelman1995bayesian}. In this approach, a global parameter set, $\theta$, influences the individual, local parameters $z$, which in turn model the observed data $D$. The relationship among these elements is encapsulated in the posterior distribution:
\begin{equation}
p(\theta,z\vert D) \propto p(D\vert z)p(z\vert\theta)p(\theta),
\end{equation}
with the data's likelihood function dependent solely on local parameters. A notable aspect of BHM is the funnel-like structure of its posterior distribution. This complex geometry can challenge the efficiency of MCMC sampling samplers, although certain reparameterizations can mitigate these difficulties~\citep{papaspiliopoulos2007general}.

\begin{figure}[ht!]
    \centering
    \includegraphics[width=7.5cm,trim={0.5cm 0.5cm 0.5cm 0.5cm},clip]{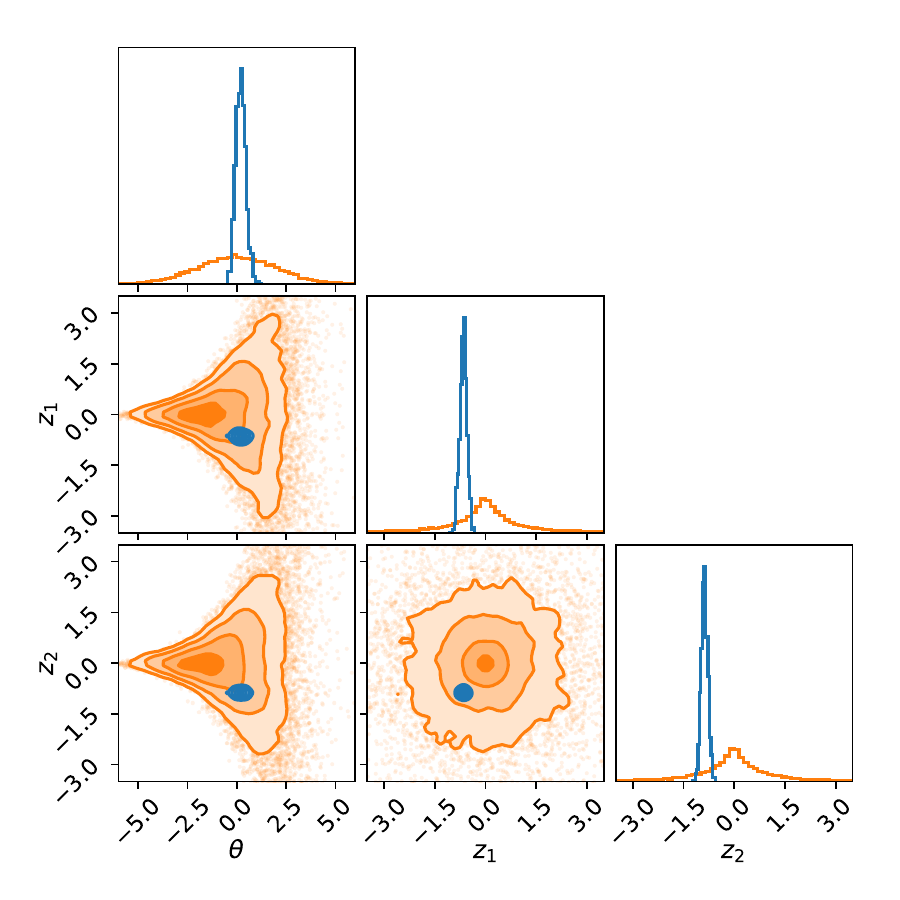}
    \caption{1D and 2D marginal posterior (\textit{blue}) and prior (\textit{orange}) distributions of the first three parameters of the Bayesian hierarchical model. }
    \label{fig:bhm}
\end{figure}

Following the concept of Neal's funnel~\citep{neal2003slice}, our model sets the global parameter distribution as $p(\theta)=\mathcal{N}(\theta\vert 0, \tau^{2})$ and the local parameter conditional distribution as $p(\mathbf{z}\vert\theta)=\mathcal{N}(\mathbf{z}\vert 0, \exp(\theta)\mathbf{I})$. We assume a normal sampling distribution for the data $p(\mathbf{D}\vert \mathbf{z}) = \mathcal{N}(\mathbf{D}\vert \mathbf{z}, \sigma^{2}\mathbf{I})$. The data for our experiment were generated under the condition $\theta = 0$ with $\tau=2$ and $\sigma=0.1$. The model considers $\theta$ as a scalar, while $\mathbf{z}$ and $\mathbf{D}$ are 30-dimensional vectors, resulting in a total of 31 parameters in the model. Fig. \ref{fig:bhm} illustrates the 1D and 2D marginal posterior contours for the first three parameters of this target. The prior-imposed funnel structure is clear in the $\theta$--$z_{i}$ contours.

\begin{figure}[ht!]
    \centering
    \includegraphics[width=16.0cm,trim={0 0.30cm 0 0.25cm},clip]{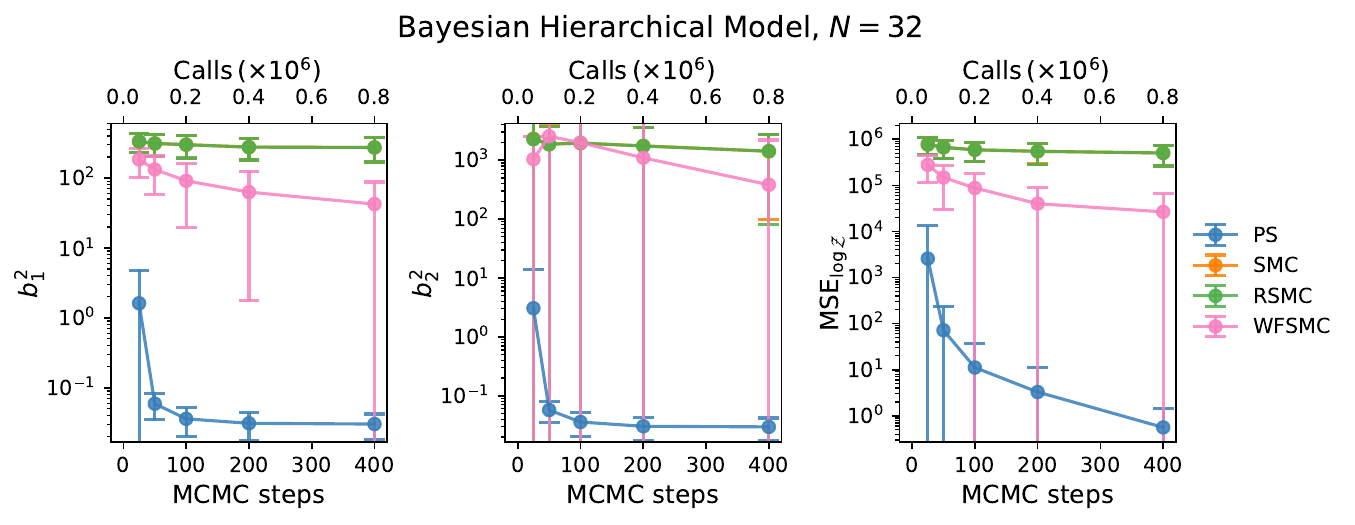}
    \vskip 0.5cm
    \includegraphics[width=16.0cm,trim={0 0.30cm 0 0.25cm},clip]{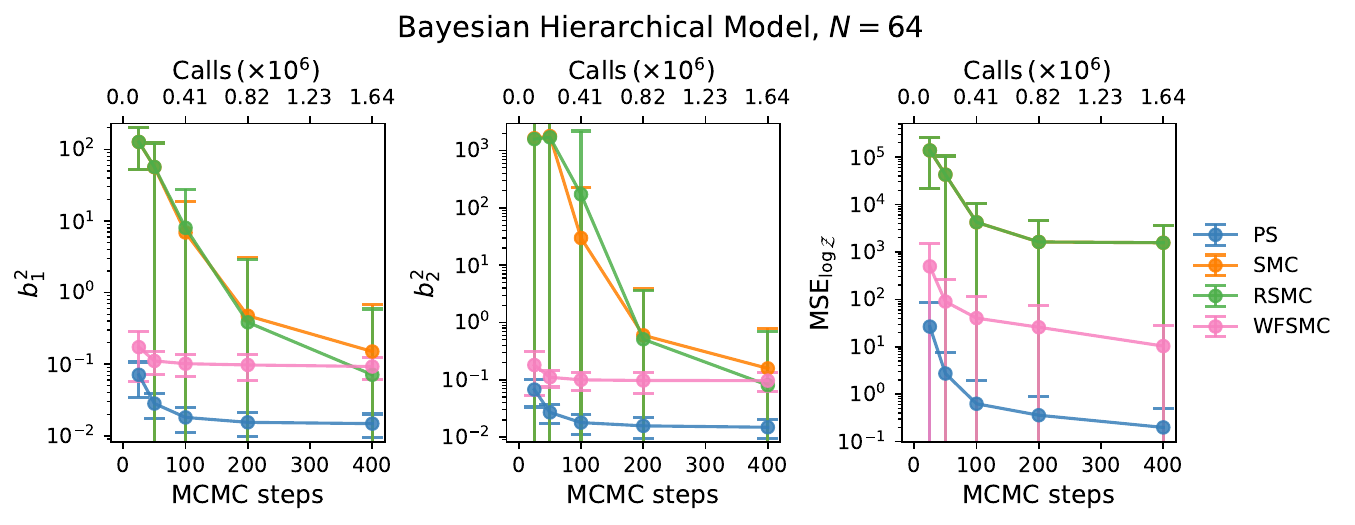}
    \vskip 0.5cm
    \includegraphics[width=16.0cm,trim={0 0.30cm 0 0.25cm},clip]{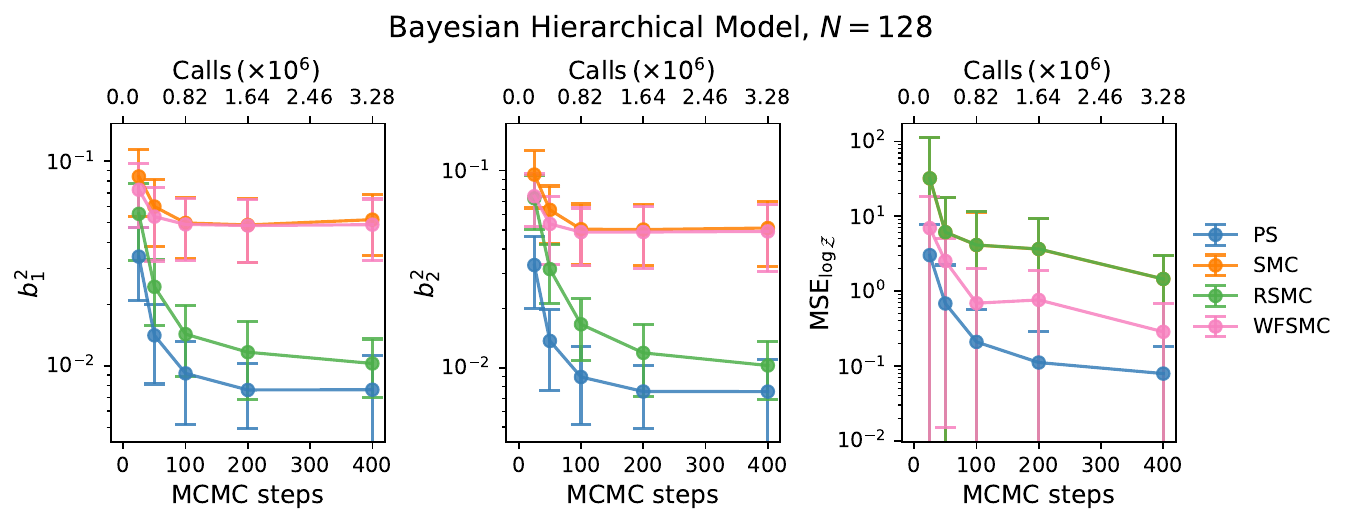}
    \caption{Sampling performance for the first (\emph{left}) and second (\emph{middle}) posterior moments and log marginal likelihood ($\textrm{MSE}_{\log \mathcal{Z}}$; \emph{right}) for the 31-dimensional Bayesian hierarchical model target using $N=32$ (\textit{top}), $N=64$ (\textit{middle}), and $N=128$ (\textit{bottom}) particles.}
    \label{fig:hier_comp}
\end{figure}

The 31-dimensional Bayesian hierarchical model benchmark underscores PS robustness and efficiency across all metrics, particularly in navigating the challenging funnel geometry of the posterior, as shown in Fig. \ref{fig:hier_comp}. PS achieves substantially lower maximum squared bias ($b_1^2$, $b_2^2$) and log marginal likelihood MSE ($\textrm{MSE}_{\log\mathcal{Z}}$) compared to SMC, RSMC, and WFSMC at all particle counts ($N \in [32, 128]$) and MCMC steps ($k \in [25, 400]$). For $N=32$, SMC and RSMC show no improvement in $b_1^2$ or $b_2^2$ with increasing $k$, performing identically, while WFSMC exhibits only minor gains—surpassing SMC/RSMC marginally but remaining far less accurate than PS. At $N=64$, SMC and RSMC gradually reduce $b_1^2$ and $b_2^2$ with higher $k$, eventually matching WFSMC at $k=400$; however, RSMC outperforms both SMC and WFSMC at this configuration. For $N=128$, RSMC surpasses SMC and WFSMC in posterior moment accuracy, though all three are outperformed by PS. Notably, WFSMC exceeds SMC/RSMC at intermediate $k$ ($\leq 200$) for $N=64$ but aligns with them at higher $k$, while PS maintains a consistent and growing advantage as $k$ increases. In $\textrm{MSE}_{\log\mathcal{Z}}$, PS achieves rapid error reduction, whereas SMC, RSMC, and WFSMC show minimal improvement even at $k=400$, often plateauing at suboptimal values. This stark contrast highlights PS’s robustness in mitigating the sampling inefficiencies induced by hierarchical funnel structures, where other methods struggle to adapt despite matched computational costs.

\section{Discussion}
\label{sec:discussion}

We introduced Persistent Sampling (PS), a novel extension of Sequential Monte Carlo (SMC) that systematically retains and reuses particles from all previous iterations to construct a growing, weighted ensemble. By leveraging multiple importance sampling and resampling from a mixture of historical distributions, PS mitigates the reliance on large particle ensembles inherent to standard SMC, thereby overcoming challenges such as limited particle diversity after resampling and mode collapse in multimodal posteriors. Crucially, PS achieves this without additional likelihood evaluations, as weights for persistent particles are computed using cached likelihood values from prior iterations. This approach not only enriches posterior approximations but also produces lower-variance estimates of the marginal likelihood—essential for robust model comparison—while enabling more efficient adaptation of transition kernels through a larger, decorrelated particle pool.

Our experiments across diverse benchmarks—ranging from high-dimensional Gaussian mixtures to hierarchical models—demonstrate that PS consistently outperforms SMC and related variants, including recycled and waste-free SMC. It achieves superior accuracy in posterior moment estimation and evidence computation, even under matched computational budgets. Notably, PS’s ability to propagate a persistent particle ensemble mitigates the need for excessively long Markov chain Monte Carlo (MCMC) runs, as resampling from a larger pool inherently reduces particle correlation. These results validate PS as a computationally scalable framework, particularly advantageous in complex inference tasks where traditional SMC struggles to balance accuracy and cost.

Despite its strong empirical performance and theoretical advantages, it is important to acknowledge the potential limitations and practical considerations of PS. While PS does not increase the number of likelihood evaluations compared to standard SMC, it does introduce computational and memory overhead. The algorithm requires re-computing weights for the entire persistent ensemble at each iteration , and although these are simple arithmetic operations, the cost can become non-trivial for runs with an extremely large number of iterations. Furthermore, storing all historical particles and their cached likelihood values leads to a memory footprint that grows linearly with the number of iterations, which could be a constraint on memory-limited systems. Another consideration is that the effectiveness of PS is inherently linked to the quality of the underlying MCMC kernel. While the growing particle pool helps adapt the kernel more robustly  and mitigates particle correlation, a fundamentally inefficient MCMC kernel that fails to explore the target space will produce a poor-quality persistent ensemble, limiting the potential gains. Finally, PS introduces a small bias in its normalizing constant estimates in exchange for a significant reduction in variance. Although we have proven the estimators are consistent  and found this bias to be negligible in practice, users in fields where unbiasedness is a strict requirement should be aware of this trade-off. These factors represent practical trade-offs rather than fundamental flaws, and our experiments demonstrate that for a wide range of challenging problems, the benefits of PS far outweigh these costs.

Furthermore, PS's computational efficiency is noteworthy. Despite maintaining a larger set of persistent particles, PS achieves its superior performance while requiring substantially fewer likelihood evaluations than SMC in most cases. Importantly, this translates into PS requiring fewer temperature levels compared to standard SMC, especially in higher-dimensional problems where the number of required temperature levels increases rapidly for SMC. This property of PS is particularly advantageous, as the computational cost of SMC can become prohibitive in such high-dimensional settings due to the need for a large number of temperature levels.  A potentially surprising aspect of PS's efficiency arises from its approximation of past samples.  While PS reweights particles as if they were drawn from a mixture of past target distributions, these particles are in fact approximate samples obtained through MCMC. The quality of this mixture approximation is inherently linked to the efficacy of the MCMC kernel employed and the number of MCMC steps performed.  Longer MCMC runs (or more efficient gradient-based MCMC kernels) would, in principle, yield samples more closely resembling independent draws from the intended mixture.  Interestingly, the inherent down-weighting of samples from earlier iterations in PS acts as a form of ``forgetting," naturally diminishing the influence of potentially less accurate early samples. This mechanism appears to be beneficial in practice, mitigating the variance of posterior expectations and marginal likelihood estimates even if the initial samples are not perfectly representative of the mixture components.  Indeed, although a rigorous analytical characterization of this approximation's impact on estimator variance remains an open challenge, our empirical findings consistently demonstrate a rapid reduction in variance for both normalizing constant and posterior expectation estimators with increasing MCMC steps in PS. This variance reduction is notably faster than observed in standard SMC, RSMC, or Waste-Free SMC.  This empirical behavior strongly suggests that treating past samples as originating from the mixture density is a practically robust and efficient approximation, possibly underpinned by a "forgetting effect" as highlighted in recent work \citep{karjalainen2023forgetting}.

Overall, our results demonstrate that PS represents a significant advancement in the field of SMC samplers, offering a more robust and efficient sampling algorithm for challenging Bayesian inference problems. By addressing the limitations of standard SMC, PS paves the way for tackling increasingly complex computational tasks arising in various scientific and engineering domains. Looking ahead, several promising research directions emerge. Investigating alternative reweighting and resampling strategies within the PS framework could further enhance its performance and numerical stability. Additionally, the interplay between PS and modern machine learning techniques, such as normalizing flows or neural networks, presents exciting opportunities for developing more sophisticated and adaptive transition kernels. Furthermore, investigating the use of different Markov transition kernels within the PS framework presents another interesting avenue. In particular, exploring gradient-based kernels could unlock PS's potential in high-dimensional settings where gradient information is available. PS is also compatible with the waste-free SMC framework~\citep{dau2022waste}. One could imagine a PS algorithm where, in each iteration, one resamples $N$ particles out of $(t-1)\times N \times k$ potential candidates, where $k$ is the length of the Markov chains.

Moreover, an intriguing connection between PS and \textit{nested sampling} (NS) has emerged. NS is a Bayesian computation technique that calculates the evidence by transforming the multi-dimensional evidence integral into a single dimension, which is then evaluated numerically using a simple sampling approach~\citep{skilling2004nested, skilling2006nested}. When setting the number of particles in PS to $N=1$ but the ESS fraction $\alpha$ to a value larger than $1$ (e.g., $\alpha=1000$), combined with a likelihood-threshold-based annealing scheme, PS appears to reduce to a Monte Carlo version of NS that does not share some of its usual limitations. These include the requirement for independent samples and the reliance on numerical trapezoid rules for evidence estimation. Exploring this avenue further could pave the way for developing new NS methods with superior theoretical properties and sampling performance.

In conclusion, PS, represents a significant methodological advancement in the field of particle-based sampling algorithms. The ability of PS to maintain a diverse and informative set of persistent particles throughout the sampling process leads to more accurate posterior characterization and marginal likelihood estimation while improving computational efficiency. By overcoming the limitations of standard SMC samplers, PS offers a powerful tool for researchers and practitioners facing challenging Bayesian inference problems across various scientific and engineering disciplines.

\section*{Acknowledgements}

This project has received funding from NSF Award Number 2311559, and from the U.S. Department of Energy, Office of Science, Office of Advanced Scientific Computing Research under Contract No. DE-AC02-05CH11231 at Lawrence Berkeley National Laboratory to enable research for Data-intensive Machine Learning and Analysis. The authors thank Victor Elvira, Christophe Andrieu, Nicolas Chopin, Michael Williams, Richard Grumitt, and Denja Vaso for helpful discussions.

\section*{Appendix A\quad Proofs}
\label{appendix}

\subsection*{A.1 Consistency of Normalizing Constant Estimates}

\begin{proof}
The consistency of the PS estimator is established by induction. We begin with the base cases $t = 1$ and $t = 2$, then proceed inductively for general $t$.

\bigskip

\noindent\textbf{Base Cases}

For $t = 1$, the result holds trivially since $\hat{\mathcal{Z}}_1 = \mathcal{Z}_1 = 1$ by definition. For $t = 2$, the estimator $\hat{\mathcal{Z}}_2$ is defined as the sample average $\frac{1}{N} \sum_{i=1}^N L(\theta_i^1)^{\beta_2}$, where the particles $\{\theta_i^1\}$ are drawn i.i.d. from the prior $\pi(\theta)$. By the Weak Law of Large Numbers (WLLN), this average converges in probability to $\mathbb{E}_{\pi}\left[\mathcal{L}(\theta)^{\beta_2}\right] = \mathcal{Z}_2$. Condition \ref{C1} ensures the integrability of $\mathcal{L}(\theta)^{\beta_2}$, validating the application of the WLLN.

\bigskip

\noindent\textbf{Inductive Hypothesis}  

Assume that for all $s < t$, the estimator $\hat{\mathcal{Z}}_s$ converges in probability to $\mathcal{Z}_s$ as $N \to \infty$. We aim to show that $\hat{\mathcal{Z}}_t \xrightarrow{p} \mathcal{Z}_t$.

\bigskip

\noindent\textbf{Inductive Step}

The estimator at iteration $t$ is given by:
\begin{equation}
\hat{\mathcal{Z}}_t = \frac{1}{t-1} \sum_{t'=1}^{t-1} \left( \frac{1}{N} \sum_{i=1}^N w_{tt'}^i \right),
\end{equation}
where the weights $w_{tt'}^i$ are defined as:
\begin{equation}
w_{tt'}^i = \frac{\mathcal{L}(\theta_i^{t'})^{\beta_t}}{\frac{1}{t-1} \sum_{s=1}^{t-1} \mathcal{L}(\theta_i^{t'})^{\beta_s} / \hat{\mathcal{Z}}_s}.
\end{equation}
To analyze the convergence of $\hat{\mathcal{Z}}_t$, we first consider the denominator in the weight expression. By the inductive hypothesis, $\hat{\mathcal{Z}}_s \xrightarrow{p} \mathcal{Z}_s$ for all $s < t$. Applying the continuous mapping theorem, the denominator $\frac{1}{t-1} \sum_{s=1}^{t-1} \mathcal{L}(\theta_i^{t'})^{\beta_s} / \hat{\mathcal{Z}}_s$ converges in probability to $\frac{1}{t-1} \sum_{s=1}^{t-1} \mathcal{L}(\theta_i^{t'})^{\beta_s} / \mathcal{Z}_s$. Under Condition \ref{C1}, the likelihood ratios $\mathcal{L}(\theta)^{\beta_t}/L(\theta)^{\beta_s}$ are bounded, ensuring the denominator is bounded away from zero. By Slutsky’s theorem, the weights $w_{tt'}^i$ therefore converge in probability to:
\begin{equation}
\bar{w}_{tt'}^i = \frac{\mathcal{L}(\theta_i^{t'})^{\beta_t}}{\frac{1}{t-1} \sum_{s=1}^{t-1} \mathcal{L}(\theta_i^{t'})^{\beta_s}/\mathcal{Z}_s}.
\end{equation}

Next, we address the distribution of the particles $\theta_i^{t'}$. By Condition \ref{C3}, the MCMC steps used to generate these particles are geometrically ergodic. As $N \to \infty$, the MCMC bias vanishes, and the particles $\theta_i^{t'}$ effectively follow the distribution $p_{t'}(\theta)$. This ensures that expectations over these particles align with the target distributions $p_{t'}(\theta)$.

We now apply the WLLN to the idealized weights $\bar{w}_{tt'}^i$. Since $\theta_i^{t'} \sim p_{t'}(\theta)$, the expectation of $\bar{w}_{tt'}^i$ is:
\begin{equation}
\mathbb{E}_{p_{t'}}\left[\bar{w}_{tt'}(\theta)\right] = \int \frac{\mathcal{L}(\theta)^{\beta_t}}{\frac{1}{t-1} \sum_{s=1}^{t-1} \mathcal{L}(\theta)^{\beta_s}/\mathcal{Z}_s} p_{t'}(\theta) d\theta = \mathcal{Z}_t.
\end{equation}
The WLLN then guarantees that $\frac{1}{N} \sum_{i=1}^N \bar{w}_{tt'}^i \xrightarrow{p} \mathcal{Z}_t$. Averaging this result over all $t' = 1, \dots, t-1$, we obtain:
\begin{equation}
\hat{\mathcal{Z}}_t = \frac{1}{t-1} \sum_{t'=1}^{t-1} \left( \frac{1}{N} \sum_{i=1}^N w_{tt'}^i \right)  \xrightarrow{p} \frac{1}{t-1} \sum_{t'=1}^{t-1} \mathcal{Z}_t = \mathcal{Z}_t.
\end{equation}
This concludes the proof. \qedhere
\end{proof}

\subsection*{A.2 Consistency of Posterior Expectations}

\begin{proof}
To establish the consistency of the PS estimator, we begin by examining the total error between our estimator and the true expectation. This error can be decomposed into two fundamental components: one arising from our use of estimated normalizing constants $\{\hat{\mathcal{Z}}_t\}$, and another from the Monte Carlo approximation itself. Formally, we can write:
\begin{equation}
    |\hat{\mu}_N(f) - \mathbb{E}_\mathcal{P}[f(\theta)]| \leq \underbrace{|\hat{\mu}_N(f) - \bar{\mu}_N(f)|}_{\text{Error from } \hat{\mathcal{Z}}_t} + \underbrace{|\bar{\mu}_N(f) - \mathbb{E}_\mathcal{P}[f(\theta)]|}_{\text{Monte Carlo error}}
\end{equation}
Here, $\hat{\mu}_N(f)$ represents our PS estimator using estimated normalizing constants, while $\bar{\mu}_N(f)$ is an idealized version using the true normalizing constants $\{\mathcal{Z}_t\}$.

Let us first address the error introduced by using estimated normalizing constants. The weights in our estimator, denoted as $W_{Tt'}^i$, depend on these constants. We can express both the estimated and true weights as:
\begin{equation}
    \hat{W}_{Tt'}^i = \frac{w_{Tt'}^i}{\hat{\mathcal{Z}}_T}, \quad \bar{W}_{Tt'}^i = \frac{\bar{w}_{Tt'}^i}{\mathcal{Z}_T}
\end{equation}

A key insight comes from Condition \ref{C1}, which ensures bounded likelihood ratios. This condition implies that our weight function exhibits Lipschitz continuity with respect to $\hat{Z}_t$:
\begin{equation}
    |\hat{W}_{Tt'}^i - \bar{W}_{Tt'}^i| \leq L \max_{1 \leq t \leq T} |\hat{\mathcal{Z}}_t - \mathcal{Z}_t|
\end{equation}
where $L$ is a constant determined by $M_{ts}$ from Condition \ref{C1} and $\epsilon$ from Condition \ref{C2}.

When we consider all particles and iterations together, this leads to:
\begin{equation}
    |\hat{\mu}_N(f) - \bar{\mu}_N(f)| \leq \|f\|_\infty \sum_{t'=1}^T \sum_{i=1}^N |\hat{W}_{Tt'}^i - \bar{W}_{Tt'}^i| \leq L T N \|f\|_\infty \max_{1 \leq t \leq T} |\hat{\mathcal{Z}}_t - \mathcal{Z}_t|
\end{equation}
This bound allows us to establish probabilistic convergence. For any $\epsilon > 0$, we can choose $\delta = \epsilon / (2 L T N \|f\|_\infty)$, giving us:
\begin{equation}
    P\left(|\hat{\mu}_N(f) - \bar{\mu}_N(f)| > \epsilon/2\right) \leq P\left(\max_{1 \leq t \leq T} |\hat{\mathcal{Z}}_t - \mathcal{Z}_t| > \delta\right)
\end{equation}
Through the union bound and Theorem 2's guarantee of $\hat{\mathcal{Z}}_t$'s consistency, we can show that this probability approaches zero as $N$ increases, proving that $|\hat{\mu}_N(f) - \bar{\mu}_N(f)| \xrightarrow{p} 0$.

Turning to the Monte Carlo error term, we examine the structure of our idealized estimator:
\begin{equation}
    \bar{\mu}_N(f) = \sum_{t'=1}^T \sum_{i=1}^N \bar{W}_{Tt'}^i f(\theta_i^{t'})
\end{equation}
where $\bar{W}_{Tt'}^i = \bar{w}_{Tt'}^i/\mathcal{Z}_T$ and $\bar{w}_{Tt'}^i = \mathcal{L}(\theta_i^{t'})^{\beta_T}/\left[\frac{1}{T-1} \sum_{s=1}^{T-1} \mathcal{L}(\theta_i^{t'})^{\beta_s}/\mathcal{Z}_s\right]$.

Condition \ref{C3} ensures that after sufficient MCMC steps, our particles approximately follow the desired distribution $p_{t'}(\theta)$. Specifically, for any bounded function $h$:
\begin{equation}
    \left|\mathbb{E}[h(\theta_i^{t'})] - \int h(\theta) p_{t'}(\theta) d\theta\right| \leq C_{t'} \|h\|_\infty \rho_{t'}^n
\end{equation}
where $\rho_{t'} < 1$, making this bias negligible for large $n$.

The weights $\bar{W}_{Tt'}^i$ serve as importance weights to correct the distribution from $p_{t'}$ to $\mathcal{P}(\theta)$. By construction:
\begin{equation}
    \mathbb{E}_{p_{t'}}[\bar{w}_{Tt'}(\theta)] = \mathcal{Z}_T
\end{equation}
and after normalization:
\begin{equation}
    \mathbb{E}_{p_{t'}}[\bar{W}_{Tt'}(\theta) f(\theta)] = \mathbb{E}_\mathcal{P}[f(\theta)]
\end{equation}

The Law of Large Numbers, combined with the i.i.d. nature of particles within iterations and Condition \ref{C5}'s boundedness of $f$, ensures that for each $t'$:
\begin{equation}
    \frac{1}{N} \sum_{i=1}^N \bar{W}_{Tt'}^i f(\theta_i^{t'}) \xrightarrow{p} \mathbb{E}_\mathcal{P}[f(\theta)]
\end{equation}
The continuous mapping theorem then gives us $\bar{\mu}_N(f) \xrightarrow{p} \mathbb{E}_\mathcal{P}[f(\theta)]$, establishing that $|\bar{\mu}_N(f) - \mathbb{E}_\mathcal{P}[f(\theta)]| \xrightarrow{p} 0$.

Finally, Slutsky's theorem allows us to combine our results. Since both error terms converge in probability to zero, we can conclude that $|\hat{\mu}_N(f) - \mathbb{E}_\mathcal{P}[f(\theta)]| \xrightarrow{p} 0$, completing our proof of the PS estimator's consistency.\qedhere
\end{proof}

\subsection*{A.3 Bias Magnitude}

\begin{proof}
We give the detailed proof in several steps. For simplicity we first explain the case $t=3$ and then indicate the extension to general $t$.

\bigskip

\noindent\textbf{The Role of the Inverse of $\hat{Z}_s$ and Taylor Expansion.}

Recall that in the PS algorithm, the weights for iteration $t$ are computed using estimates from previous iterations. In particular, for a given previous iteration $t'$ (with $1\le t'<t$) the weight for the $i$th particle is given by
\begin{equation}
w_{tt'}^i = \frac{\mathcal{L}(\theta_i^{t'})^{\beta_t}}{\displaystyle \frac{1}{t-1}\sum_{s=1}^{t-1} \frac{\mathcal{L}(\theta_i^{t'})^{\beta_s}}{\hat{\mathcal{Z}}_s}}.
\end{equation}
The nonlinearity in this expression arises because of the inverses $\hat{\mathcal{Z}}_s^{-1}$. Even if $\hat{\mathcal{Z}}_s$ is a consistent estimator of $\mathcal{Z}_s$, the function $x\mapsto 1/x$ is convex; by Jensen's inequality, we do not have $\E[1/\hat{\mathcal{Z}}_s]=1/\mathcal{Z}_s$ exactly. 

To quantify this bias we use a second-order Taylor expansion of the function $f(x)=1/x$ around the true value $x=\mathcal{Z}_s$. Writing
\begin{equation}
\hat{\mathcal{Z}}_s = \mathcal{Z}_s + \Delta_s, \quad \text{with } \Delta_s = \hat{\mathcal{Z}}_s - \mathcal{Z}_s,
\end{equation}
we expand
\begin{equation}
\frac{1}{\hat{\mathcal{Z}}_s} = \frac{1}{\mathcal{Z}_s+\Delta_s} = \frac{1}{\mathcal{Z}_s} - \frac{\Delta_s}{\mathcal{Z}_s^2} + \frac{\Delta_s^2}{\mathcal{Z}_s^3} + \mathcal{O}(\Delta_s^3).
\end{equation}
Taking expectation and noting that $\E[\Delta_s]=\E[\hat{\mathcal{Z}}_s] - \mathcal{Z}_s$ (which is of lower order) and $\E[\Delta_s^2]=\Var(\hat{\mathcal{Z}}_s)$, we obtain
\begin{equation}
\E\left[\frac{1}{\hat{\mathcal{Z}}_s}\right] = \frac{1}{\mathcal{Z}_s} + \frac{\Var(\hat{\mathcal{Z}}_s)}{\mathcal{Z}_s^3} + \mathcal{O}\left(\frac{1}{N^2}\right).
\end{equation}
Since by the Monte Carlo nature of the estimator, we typically have
\begin{equation}
\Var(\hat{\mathcal{Z}}_s) = \frac{\sigma_s^2}{N},
\end{equation}
the extra term in the expectation is of order $1/N$.

\bigskip

\noindent\textbf{Analysis for the Case $t=3$.}

For $t=3$ the estimator is
\begin{equation}
\hat{\mathcal{Z}}_3 = \frac{1}{2}\sum_{t'=1}^{2}\left(\frac{1}{N}\sum_{i=1}^N w_{3t'}^i\right).
\end{equation}
Focus on the contribution from particles generated at $t'=2$. (A similar argument holds for $t'=1$; in many practical cases, the first iteration is trivial since $\hat{\mathcal{Z}}_1=\mathcal{Z}_1=1$ when sampling from the normalized prior.)

For $t'=2$, the weight is
\begin{equation}
w_{32}^i = \frac{\mathcal{L}(\theta_i^{2})^{\beta_3}}{\displaystyle \frac{1}{2}\Bigl(\mathcal{L}(\theta_i^{2})^{\beta_1}\frac{1}{\hat{\mathcal{Z}}_1} + \mathcal{L}(\theta_i^{2})^{\beta_2}\frac{1}{\hat{\mathcal{Z}}_2}\Bigr)}.
\end{equation}
Since $\hat{\mathcal{Z}}_1=\mathcal{Z}_1$ exactly, we write the denominator as
\begin{equation}
D_i = \frac{1}{2}\left(\mathcal{L}(\theta_i^{2})^{\beta_1} + \mathcal{L}(\theta_i^{2})^{\beta_2}\frac{1}{\hat{\mathcal{Z}}_2}\right).
\end{equation}
Now, substitute the Taylor expansion for $1/\hat{\mathcal{Z}}_2$:
\begin{equation}
\frac{1}{\hat{\mathcal{Z}}_2} = \frac{1}{\mathcal{Z}_2} + \frac{\Var(\hat{\mathcal{Z}}_2)}{\mathcal{Z}_2^3} + \mathcal{O}\left(\frac{1}{N^2}\right).
\end{equation}
Thus, the denominator becomes
\begin{equation}
D_i = \frac{1}{2}\left(\mathcal{L}(\theta_i^{2})^{\beta_1} + \mathcal{L}(\theta_i^{2})^{\beta_2}\left[\frac{1}{\mathcal{Z}_2} + \frac{\Var(\hat{\mathcal{Z}}_2)}{\mathcal{Z}_2^3}\right]\right) + \mathcal{O}\left(\frac{1}{N^2}\right).
\end{equation}
Define the ideal (or \emph{true}) weight that one would have if the true normalizing constant were used:
\begin{equation}
\bar{w}_{32}^i = \frac{\mathcal{L}(\theta_i^{2})^{\beta_3}}{\displaystyle \frac{1}{2}\left(\mathcal{L}(\theta_i^{2})^{\beta_1} + \mathcal{L}(\theta_i^{2})^{\beta_2}\frac{1}{\mathcal{Z}_2}\right)}.
\end{equation}
Then we can write
\begin{equation}
w_{32}^i = \bar{w}_{32}^i \cdot \left(1 - \frac{\mathcal{L}(\theta_i^{2})^{\beta_2}}{\mathcal{L}(\theta_i^{2})^{\beta_1} + \mathcal{L}(\theta_i^{2})^{\beta_2}\frac{1}{\mathcal{Z}_2}} \cdot \frac{\Var(\hat{\mathcal{Z}}_2)}{\mathcal{Z}_2^2} \right) + \mathcal{O}\left(\frac{1}{N^2}\right).
\end{equation}
Taking the expectation with respect to the particle distribution at iteration $2$ (and using the fact that the function
\begin{equation}
\theta \mapsto \frac{\mathcal{L}(\theta)^{\beta_2}}{\mathcal{L}(\theta)^{\beta_1} + \mathcal{L}(\theta)^{\beta_2}\frac{1}{\mathcal{Z}_2}}
\end{equation}
is bounded by Condition \ref{C1}), we obtain
\begin{equation}
\E[w_{32}^i] = \E\left[\bar{w}_{32}^i\right] \left(1 + \frac{1}{2}\frac{\Var(\hat{\mathcal{Z}}_2)}{\mathcal{Z}_2^2} + \mathcal{O}\left(\frac{1}{N^2}\right)\right).
\end{equation}
But by the design of the weights $\bar{w}_{32}^i$, one has
\begin{equation}
\E\left[\bar{w}_{32}^i\right] = \mathcal{Z}_3.
\end{equation}
Thus,
\begin{equation}
\E[w_{32}^i] = \mathcal{Z}_3\left(1 + \frac{\Var(\hat{\mathcal{Z}}_2)}{2\mathcal{Z}_2^2} + \mathcal{O}\left(\frac{1}{N^2}\right)\right).
\end{equation}
Since $\Var(\hat{\mathcal{Z}}_2)=\sigma_2^2/N$, we conclude that
\begin{equation}
\left|\frac{\E[w_{32}^i]}{\mathcal{Z}_3} - 1\right| \le \frac{\sigma_2^2}{2N\mathcal{Z}_2^2} + \mathcal{O}\left(\frac{1}{N^2}\right).
\end{equation}
Averaging over both contributions (from $t'=1$ and $t'=2$) in the definition of $\hat{\mathcal{Z}}_3$ does not change the order of the bias. In other words, for $t=3$,
\begin{equation}
\left|\frac{\E[\hat{\mathcal{Z}}_3]}{Z_3} - 1\right| \le \frac{C_3}{N} + \mathcal{O}\left(\frac{1}{N^2}\right),
\end{equation}
with $C_3$ proportional to $\sigma_2^2/\mathcal{Z}_2^2$.

\bigskip

\noindent\textbf{Extension to General $t$.}

For a general iteration $t\ge3$, the estimator is
\begin{equation}
\hat{\mathcal{Z}}_t = \frac{1}{t-1}\sum_{t'=1}^{t-1}\left(\frac{1}{N}\sum_{i=1}^N w_{tt'}^i\right),
\end{equation}
with weights given by
\begin{equation}
w_{tt'}^i = \frac{\mathcal{L}(\theta_i^{t'})^{\beta_t}}{\displaystyle \frac{1}{t-1}\sum_{s=1}^{t-1} \frac{\mathcal{L}(\theta_i^{t'})^{\beta_s}}{\hat{\mathcal{Z}}_s}}.
\end{equation}
Repeating the analysis as in the $t=3$ case for each previous iteration $s$ (with $s=1,\ldots,t-1$) one obtains a similar Taylor expansion for each $\hat{\mathcal{Z}}_s$. In every term the error introduced by using $\hat{\mathcal{Z}}_s$ in place of $\mathcal{Z}_s$ appears as
\begin{equation}
\frac{\Var(\hat{\mathcal{Z}}_s)}{Z_s^3} \propto \frac{1}{N},
\end{equation}
and when averaged over the $t-1$ iterations, the overall bias satisfies
\begin{equation}
\left|\frac{\E[\hat{\mathcal{Z}}_t]}{\mathcal{Z}_t} - 1\right| \le \frac{t-1}{2N} \sum_{s=2}^{t-1}\frac{\sigma_s^2}{\mathcal{Z}_s^2} + \mathcal{O}\left(\frac{1}{N^2}\right).
\end{equation}
Defining
\begin{equation}
    C_t = \frac{t-1}{2}\sum_{s=2}^{t-1}\frac{\sigma_s^2}{\mathcal{Z}_s^2},
\end{equation}
we have proved the claim. \qedhere

\end{proof}

\bibliographystyle{unsrtnat}
\bibliography{references}

\end{document}